%% file: main.tex
\documentclass[10pt,twocolumn,letterpaper]{article}
\usepackage{iccv}
\usepackage{times}
\usepackage{epsfig}
\usepackage{graphicx}
\usepackage{amsmath}
\usepackage{amssymb}
\usepackage{bm}
\usepackage{color}          
\usepackage{microtype}      
\usepackage{dcolumn}        
\usepackage[pagebackref=true,breaklinks=true,letterpaper=true,colorlinks,bookmarks=false]{hyperref}
\usepackage{url}            
\usepackage{booktabs}       
\usepackage{amsfonts}       
\usepackage{nicefrac}       

\usepackage{enumitem}
\usepackage{multirow}
\usepackage{float}
\usepackage{algorithm}
\usepackage[noend]{algpseudocode}
\usepackage{amsthm}
\usepackage{subfiles}

\DeclareMathOperator*{\argmin}{argmin}
\newtheorem{theorem}{Theorem}
\newtheorem{corollary}{Corollary}[theorem]
\newtheorem{lemma}[theorem]{Lemma}

\iccvfinalcopy 

\def\iccvPaperID{762} 

\ificcvfinal\fi
\begin{document}

\title{Adversarial Domain Randomization}

\author{Rawal Khirodkar \hspace{2cm} Kris M. Kitani \\
Carnegie Mellon University\\
{\tt\small \{rkhirodk,kkitani\}@cs.cmu.edu}
}

\maketitle

\begin{abstract}
Domain Randomization (DR) is known to require a significant amount of training data for good performance \cite{khirodkar2018domain,tremblay2018training}. We argue that this is due to DR's strategy of random data generation using a uniform distribution over simulation parameters, as a result, DR often generates samples which are uninformative for the learner. In this work, we theoretically analyze DR using ideas from multi-source domain adaptation. Based on our findings, we propose Adversarial Domain Randomization (ADR) as an efficient variant of DR which generates adversarial samples with respect to the learner during training. We implement ADR as a policy whose action space is the quantized simulation parameter space. At each iteration, the policy's action generates labeled data and the reward is set as negative of learner's loss on this data. As a result, we observe ADR frequently generates novel samples for the learner like truncated and occluded objects for object detection and confusing classes for image classification. We perform evaluations on datasets like CLEVR, Syn2Real, and VIRAT for various tasks where we demonstrate that ADR outperforms DR by generating fewer data samples.
\end{abstract}

\vspace{-6mm}
\section{Introduction}
\subfile{sections/introduction}

\section{Related Work}
\subfile{sections/related_work}

\section{Theoretical Analysis of DR}
\label{section:thoery_dr}
\subfile{sections/theoretical_analysis}

\section{Adversarial Domain Randomization}
\label{section:adr}
\subfile{sections/method}

\section{Experiments}
\subfile{sections/experiments}

\section{Conclusion}

DR is a powerful method that bridges the gap between real and synthetic data. There is a need to analyze such an important technique, our work is the first step in this direction. We theoretically show that DR is superior to synthetic data generation without randomization. We also identify DR's requirement of a lot of data for generalization. As an alternative, we proposed ADR, which generates adversarial samples with respect to the learner during training. Our evaluations show that ADR outperforms DR using less data for image classification and object detection on real datasets.

\newpage
{\small
\bibliographystyle{ieee}
\bibliography{references}
}

\end{document}


\title{Adversarial Domain Randomization}

\author{Rawal Khirodkar \hspace{2cm} Donghyun Yoo \hspace{2cm} Kris M. Kitani \\
Carnegie Mellon University\\
{\tt\small \{rkhirodk,donghyuy,kkitani\}@cs.cmu.edu}
}

\maketitle







\begin{lemma}{(Based on Lemma 3 \cite{ben2010theory})}
\label{lemma:divergence}
For any hypotheses $h_1, h_2 \in \mathcal{H}$,

\vspace{-5mm}
\begin{align*}
|\epsilon_\alpha(h_1, h_2) - \epsilon_T(h_1, h_2)| \leq \frac{1}{2} d_{\mathcal{H}\Delta\mathcal{H}} (\mathcal{D}_\alpha, \mathcal{D}_T) 
\end{align*}
\end{lemma}

\begin{proof}
By the definition of $\mathcal{H}\Delta\mathcal{H}$ divergence,
$d_{\mathcal{H}\Delta\mathcal{H}} (\mathcal{D}_\alpha, \mathcal{D}_T)$
\begin{align*}
 &=   2 \sup_{h, h' \in \mathcal{H}} | \text{Pr}_{x \sim \mathcal{D}_\alpha}[h(x) \neq h'(x)] - \text{Pr}_{x \sim \mathcal{D}_T}[h(x) \neq h'(x)]|\\
 &=  2 \sup_{h, h' \in \mathcal{H}} | \epsilon_\alpha(h, h') - \epsilon_T(h, h')|\\
 &\geq 2 | \epsilon_\alpha(h_1, h_2) - \epsilon_T(h_1, h_2)|
\end{align*}
\end{proof}

\begin{lemma}{(Based on Th.5 \cite{ben2010theory})}
\label{lemma:true_alpha_target}
For fixed $h \in \mathcal{H}$ and $\alpha$, if $\gamma_\alpha = \min_h \{ \epsilon_T(h) + \epsilon_\alpha(h) \} $, then

\vspace{-5mm}
\begin{align*}
|\epsilon_\alpha(h) - \epsilon_T(h)| \leq \gamma_\alpha + \frac{1}{2} d_{\mathcal{H}\Delta\mathcal{H}} (\mathcal{D}_\alpha, \mathcal{D}_T) 
\end{align*}
\end{lemma}

\begin{proof}
$|\epsilon_\alpha(h) - \epsilon_T(h)|$
\begin{align*}
\leq & |\epsilon_\alpha(h) - \epsilon_\alpha(h, h^*)| + |\epsilon_\alpha(h, h^*) - \epsilon_T(h, h^*)|\\
+ & |\epsilon_T(h, h^*) - \epsilon_T(h)|\\
\leq & \epsilon_\alpha(h^*) + |\epsilon_\alpha(h, h^*) - \epsilon_T(h, h^*)| + \epsilon_T(h^*)\\
\leq & \gamma_\alpha + \frac{1}{2} d_{\mathcal{H}\Delta\mathcal{H}} (\mathcal{D}_\alpha, \mathcal{D}_T)
\end{align*}
\end{proof}

\begin{lemma} For $i\in \{1,2,..N\}$, let $\alpha_i = [0,.. \underset{i^{\text{th}}}{1}.. 0] $, $\gamma_i = \min_h \{\epsilon_T(h) + \epsilon_i(h) \}$, $\bar{\alpha} = [\frac{1}{N}, \frac{1}{N}..., \frac{1}{N}] $, $\bar{\gamma} = \min_h \{ \epsilon_T(h) + \frac{1}{N} \sum_{i=1}^{N} \epsilon_i(h) \}$, then
\label{lemma:comparison}

\vspace{-5mm}
\begin{align*}
\frac{1}{N} \sum_{i=1}^{N} \Big( 2\gamma_i + d_{\mathcal{H}\Delta\mathcal{H}} (\mathcal{D}_i, \mathcal{D}_T)\Big) \geq 2\bar{\gamma} + d_{\mathcal{H}\Delta\mathcal{H}} (\mathcal{D}_{\bar{\alpha}}, \mathcal{D}_T)
\end{align*}
\end{lemma}

\begin{proof}
Let the input source distributions $\mathcal{D}_1, \mathcal{D}_2,\ldots \mathcal{D}_N$ be represented by density functions $p_1, p_2, \ldots p_N$. Similarly, let the input target distribution $\mathcal{D}_T$ be represented by the density function $q$. We can now represent each ith source domain by two functions $p_i$ and $f_i$.  

Consider an arbitrary source domain $S$ represented by functions $p$ and $f$. We show that $\gamma = \min_h \{ \epsilon_T(h) + \epsilon_S(h) \}$ and $d_{\mathcal{H}\Delta\mathcal{H}} (\mathcal{D}_S, \mathcal{D}_T)$ is convex in $p$ and $f$.

Clearly, $\epsilon_S(h)$ is convex in $p$ and $f$. Therefore, $\gamma$ is convex in $p$ and $f$. We now show that the $\mathcal{H}-$ divergence $d_{\mathcal{H}}(\mathcal{D}_S, \mathcal{D}_T)$ is convex in $p$ for all $\mathcal{H}$. $d_{\mathcal{H}\Delta\mathcal{H}} = d_{\mathcal{H'}}$ such that $\mathcal{H}' = \mathcal{H}\Delta\mathcal{H}$.

$\forall \mathcal{H}$, $d_\mathcal{H} (p, q)$ is convex in $\forall q$, $\forall p_1, p_2$, $\lambda \in [0,1]$.  \\
$d_\mathcal{H}(\lambda p_1 + (1 - \lambda) p_2, q)$
\begin{align*}
     = & \text{sup}_{A \in \mathcal{H}^{-1}} |(\lambda p_1 + (1 - \lambda) p_2)(A) - q(A)|\\
     = &  |(\lambda p_1 + (1 - \lambda) p_2)(A^*) - q(A^*)|\\
     \leq &  \lambda |p_1(A^*) - q(A^*)| + (1 - \lambda)|p_2(A^*) -q(A^*)|\\
     \leq &  \lambda \text{sup}_{A \in \mathcal{H}^{-1}}|p_1(A) - q(A)| + (1 - \lambda)\text{sup}_{A \in \mathcal{H}^{-1}}|p_2(A) -q(A)|\\
     \leq &  \lambda d_\mathcal{H}(p_1, q) + (1 - \lambda)d_\mathcal{H}(p_2, q)
\end{align*}

The lemma follows from using Jensen' inquality $f(\mathbb{E}[X]) \leq \mathbb{E}[f(X)]$ where $f = \gamma_S + d_{\mathcal{H}\Delta\mathcal{H}}(\mathcal{D}_S, \mathcal{D}_T)$ is a convex function and $X$ is random variable represents choice over $N$ source domains.
\end{proof}

\begin{theorem}
\label{theorem:my} {(Based on Th.5 \cite{ben2010theory})}
Consider the optimal hypothesis on target domain $h^*_T = \argmin_h \epsilon_T(h)$ and on $\alpha$-combination of source domains $h^*_\alpha = \argmin_h \epsilon_\alpha(h)$. If $\gamma_\alpha = \min_h \{ \epsilon_T(h) + \epsilon_\alpha(h) \}$, then
\begin{align*}
\epsilon_{T}(h^*_\alpha) & \leq \epsilon_T(h^*_T) + 2\gamma_\alpha + d_{\mathcal{H}\Delta\mathcal{H}} (\mathcal{D}_{\alpha}, \mathcal{D}_T))
\end{align*}
\end{theorem}

\begin{proof}
\begin{align*}
|\epsilon_{\alpha}(h^*_T) - \epsilon_T(h^*_T)| & \leq  \gamma_{\alpha} + \frac{1}{2} d_{\mathcal{H}\Delta\mathcal{H}} (\mathcal{D}_{\alpha}, \mathcal{D}_T) && \text{L\ref{lemma:true_alpha_target}}\\
\epsilon_{\alpha}(h^*_T) & \leq \epsilon_T(h^*_T) +  \gamma_{\alpha} + \frac{1}{2} d_{\mathcal{H}\Delta\mathcal{H}} (\mathcal{D}_\alpha, \mathcal{D}_T) \\
\epsilon_{\alpha}(h^*_\alpha) & \leq \epsilon_T(h^*_T) +  \gamma_{\alpha} + \frac{1}{2} d_{\mathcal{H}\Delta\mathcal{H}} (\mathcal{D}_\alpha, \mathcal{D}_T)\\
& \because \epsilon_\alpha(h^*_\alpha) \leq \epsilon_\alpha(h^*_T)\\
|\epsilon_{\alpha}(h^*_\alpha) - \epsilon_T(h^*_\alpha)| & \leq  \gamma_{\alpha} + \frac{1}{2} d_{\mathcal{H}\Delta\mathcal{H}} (\mathcal{D}_{\alpha}, \mathcal{D}_T) && \text{L\ref{lemma:true_alpha_target}}\\
\epsilon_T(h^*_\alpha) & \leq \epsilon_{\alpha}(h^*_\alpha) + \gamma_{\alpha} + \frac{1}{2} d_{\mathcal{H}\Delta\mathcal{H}} (\mathcal{D}_{\alpha}, \mathcal{D}_T) \\
\epsilon_T(h^*_\alpha) & \leq \epsilon_T(h^*_T) + 2\gamma_{\alpha} + d_{\mathcal{H}\Delta\mathcal{H}} (\mathcal{D}_{\alpha}, \mathcal{D}_T)
\end{align*}
\end{proof}

\begin{corollary}
The generalization error bound for an equal weighted combination of source domains (DR) is smaller than the
expected generalization error bound of a single source domain (expectation over an uniform choice of source domain). 
\label{corollary:my}
\end{corollary}

\begin{proof}
For $i \in \{1, 2, .. N\} $, let $h^*_i = \argmin_h \epsilon_i(h)$ then using $\alpha = \alpha_i = [0,.. \underset{i^{\text{th}}}{1}.. 0]$ in Th.\ref{theorem:my}, we have 
\begin{align*}
    \epsilon_{T}(h^*_i) & \leq \epsilon_T(h^*_T) + 2\gamma_i + d_{\mathcal{H}\Delta\mathcal{H}} (\mathcal{D}_i, \mathcal{D}_T))\\
    \mathbb{E}_{i \sim \mathcal{U}_{N}}[\epsilon_{T}(h^*_i)] & \leq \mathbb{E}_{i \sim \mathcal{U}_{N}}[\epsilon_T(h^*_T) + 2\gamma_i + d_{\mathcal{H}\Delta\mathcal{H}} (\mathcal{D}_i, \mathcal{D}_T)]\\
    \frac{1}{N} \sum_{i=1}^{N} \epsilon_{T}(h^*_i) & \leq \epsilon_T(h^*_T) + \frac{1}{N} \sum_{i=1}^{N} \Big(2\gamma_i + d_{\mathcal{H}\Delta\mathcal{H}} (\mathcal{D}_i, \mathcal{D}_T) \Big)
\end{align*}

Similarly, for an equal weighted combination of source domains, let $\bar{h}^* = \frac{1}{N}\argmin_h \sum_{i=1}^{N} \epsilon_i(h)$ then using $\alpha = \bar{\alpha} = [\frac{1}{N},.. \frac{1}{N}]$ in Th. \ref{theorem:my}, we have
\begin{align*}
    \epsilon_{T}(\bar{h}^*) & \leq \epsilon_T(h^*_T) + 2\bar{\gamma} + d_{\mathcal{H}\Delta\mathcal{H}} (\mathcal{D}_{\bar{\alpha}}, \mathcal{D}_T)
\end{align*}

The relationship between the upper bounds follows from L\ref{lemma:comparison}.
\end{proof}

\section{CLEVR: Color Classification}

\begin{figure}[H]
 \begin{center}
  \includegraphics[width=\linewidth]{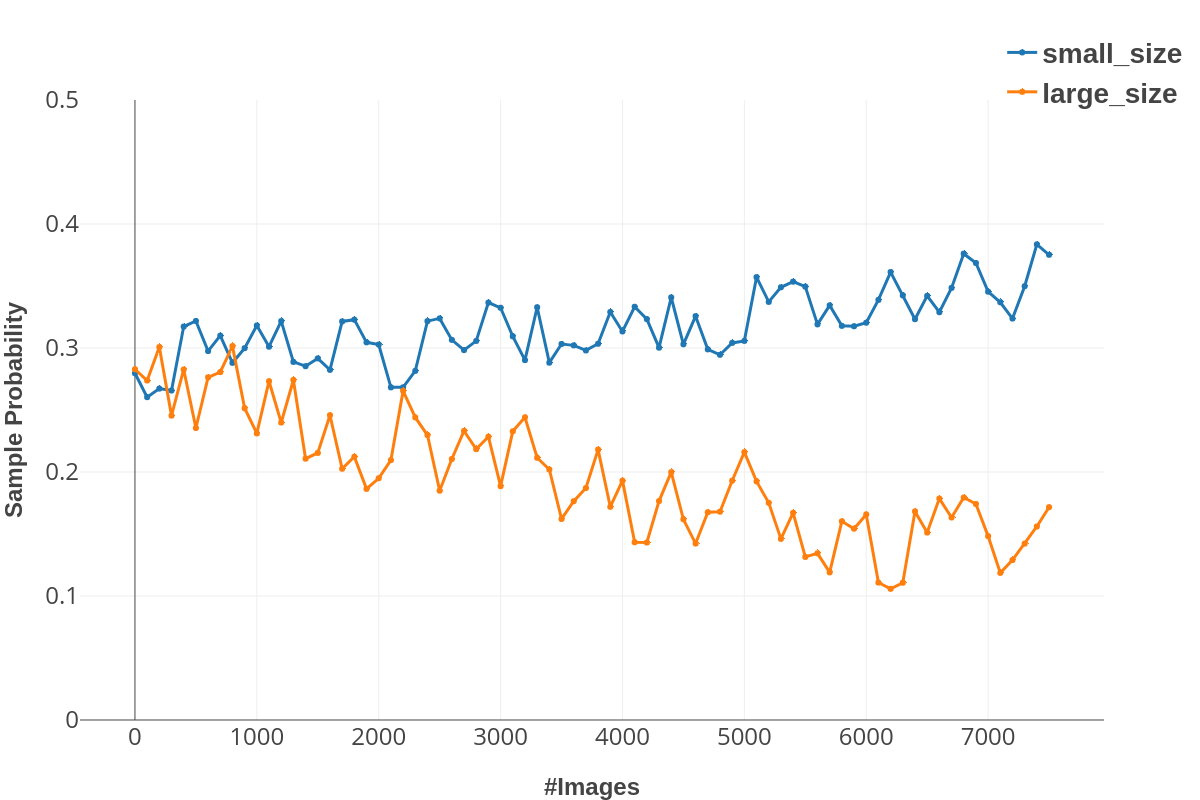}
 \end{center}
  \caption{ADR sample probability over size dimension according to $\pi_\omega$ averaged over 10 runs. small-size are the two smallest sizes and large-size are the two largest sizes.  }
 \label{fig:clevr}
 \end{figure}

Figure \ref{fig:clevr} shows the probability of generation over the most critical rendering dimension for this toy problem i.e size. ADR's policy $\pi_\omega$ effectively focuses on harder (smaller) objects to achieve better performance on target data. Each iteration corresponds to


\section{Syn2Real: Object Classification}
\begin{figure}[H]
 \begin{center}
  \includegraphics[width=\linewidth]{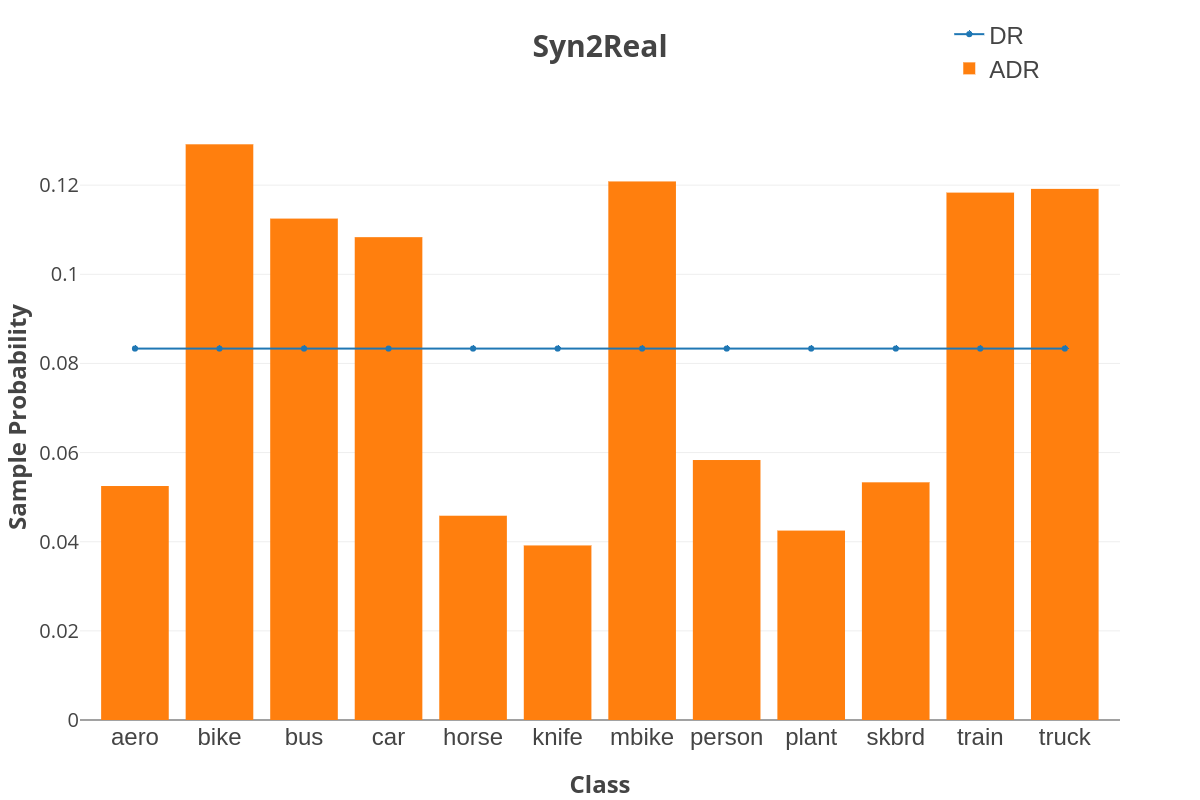}
 \end{center}
   \caption{ADR $\pi_\omega$ after generating 100k images on Syn2Real Image Classification task. ADR increases the sample generation probability of confusing classes like \{bike, mbike\} and \{bus, car, truck, train\} whereas easier classes like \{knife, plant, aero\} see a decrease in generation probability.}
 \label{fig:syn2real}
 \end{figure}

Figure \ref{fig:syn2real} visualizes sample generation probability over all the 12 classes after generating 100k images. Our proposed method focuses on confusing group of classes like \{bike, motorbike\} and \{bus, car, train, truck \} rather than generating visually distinct classes like \{aero, horse, knife, person, plant, skbrd\}.

\section{VIRAT: Simulator}
\begin{figure}[H]
 \begin{center}
  \includegraphics[width=\linewidth]{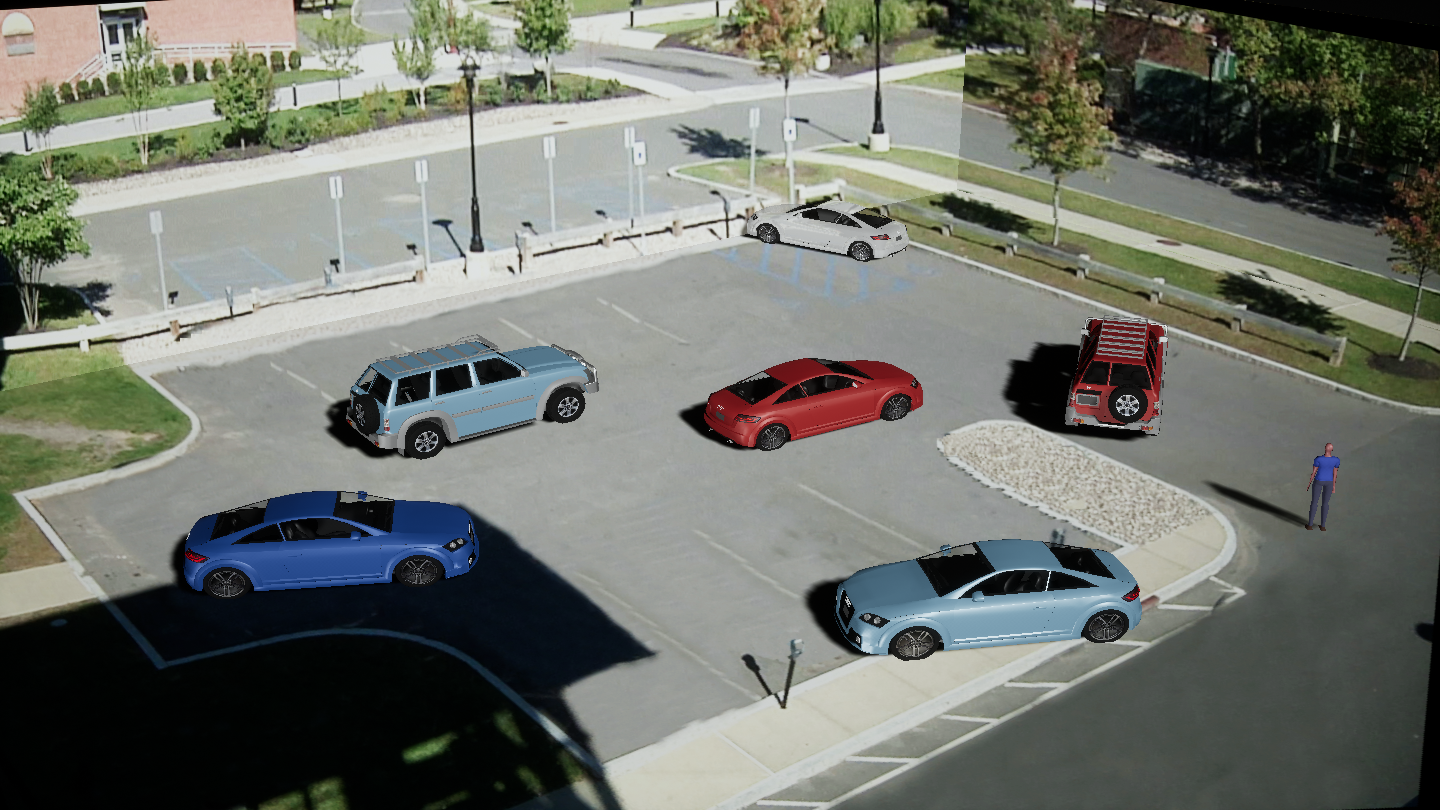}
 \end{center}
   \caption{}
 \label{fig:car_scene}
 \end{figure}

VIRAT is a public surveillance dataset, for each surveillance scene the simulator mimics the 3D geometry of the scene and the camera parameters as shown in figure \ref{fig:car_scene} using UnrealEngine 4. CAD models of objects like car, person are then placed in the 3D scene and a 2D image is captured from the camera perspective as shown in \ref{fig:car_scene}

 \begin{figure}[H]
 \begin{center}
  \includegraphics[width=\linewidth]{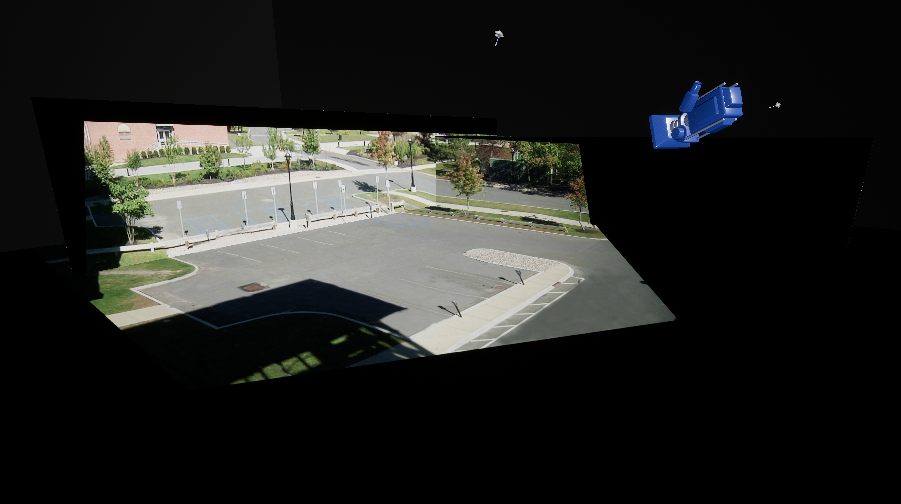}
 \end{center}
   \caption{}
 \label{fig:3d_scene}
 \end{figure}

\begin{figure}[H]
 \begin{center}
  \includegraphics[width=\linewidth]{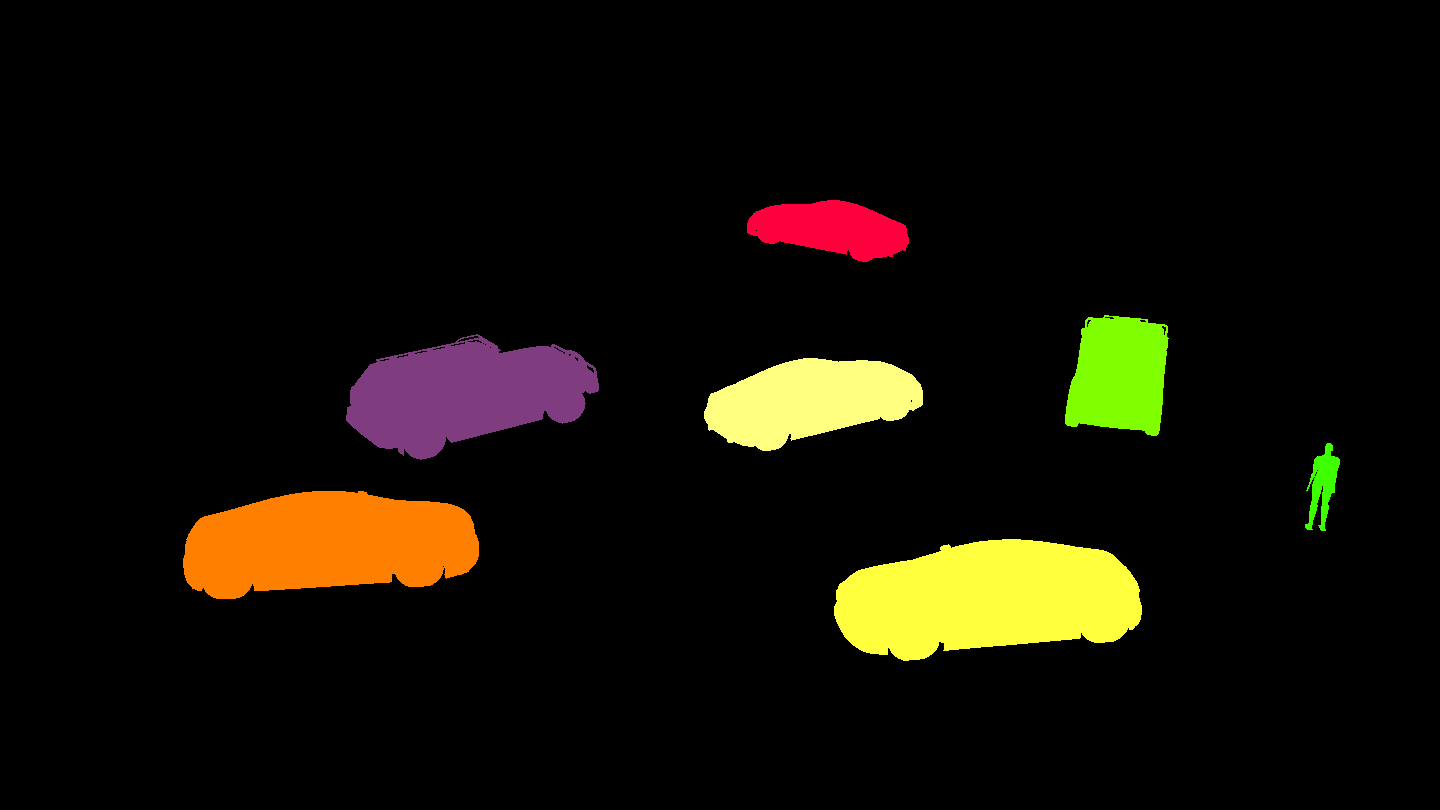}
 \end{center}
   \caption{}
 \label{fig:seg_scene}
 \end{figure}

\begin{figure}[H]
 \begin{center}
  \includegraphics[width=\linewidth]{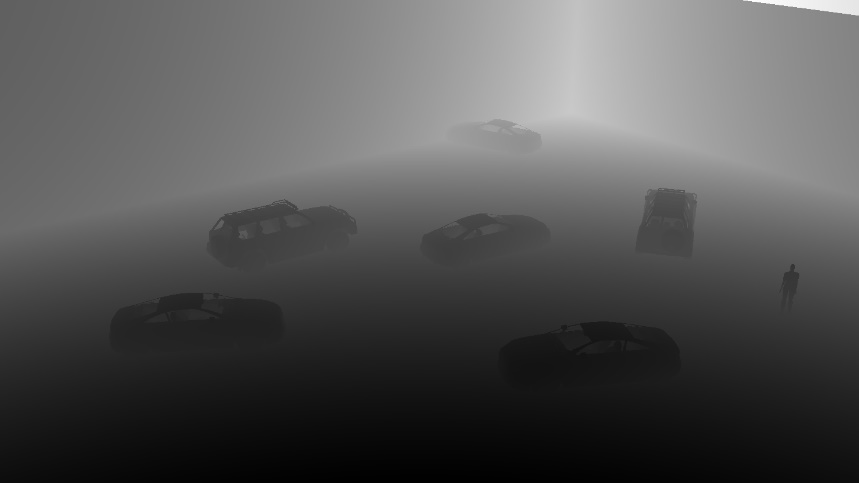}
 \end{center}
   \caption{}
 \label{fig:depth_scene}
 \end{figure}

 The simulator then uses constant shaders to extract instance segmentation maps and ray tracing to extract depth maps of the configuration as shown in figure \ref{fig:seg_scene} and \ref{fig:depth_scene}.
 
\begin{figure}[H]
 \begin{center}
  \includegraphics[width=\linewidth]{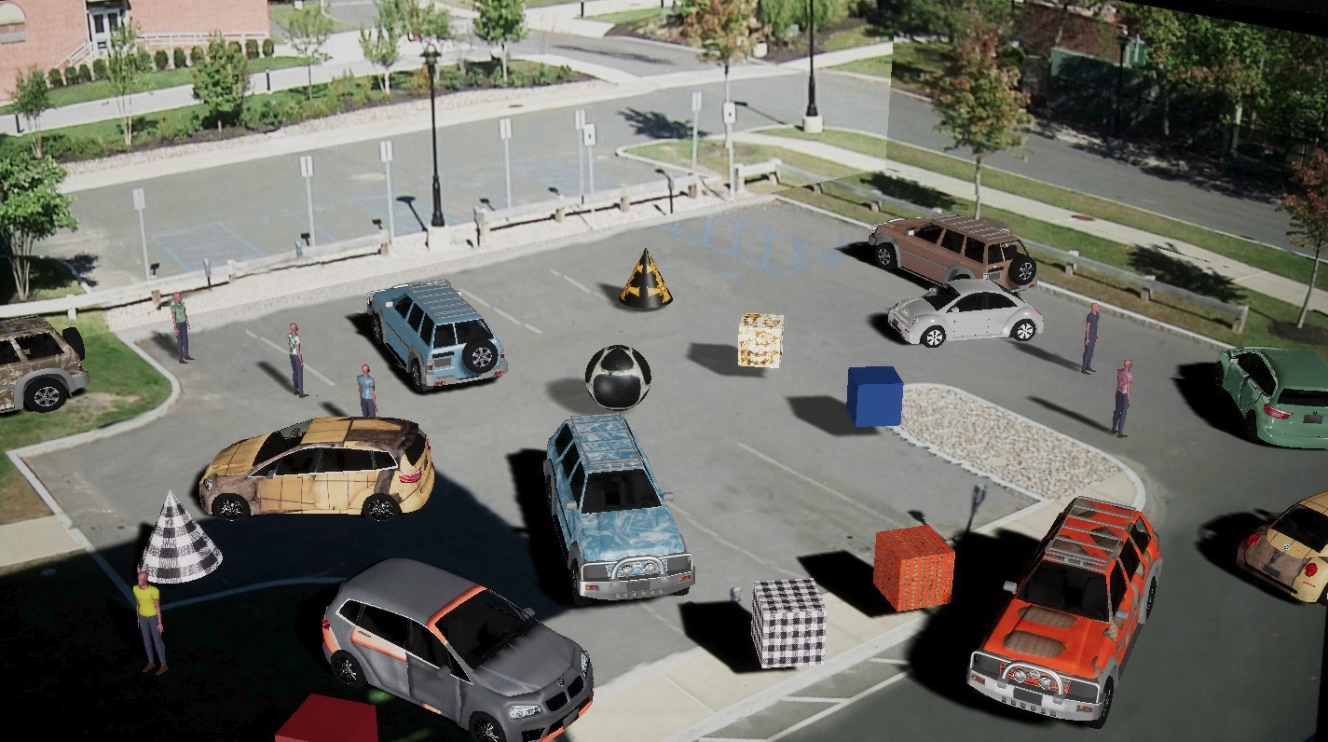}
 \end{center}
   \caption{}
 \label{fig:dr_scene}
 \end{figure}

 However, using on the synthetic data results in poor performance over the target data, we therefore perform randomization over the rendering by varying the lighting conditions, textures, dimensions of the objects placed in the scene as shown in figure \ref{fig:dr_scene}.

\section{VIRAT: Depth Estimation}
In this section, we demonstrate the utility of our approach on a challenging use-case. We perform monocular depth estimation on VIRAT dataset using synthetic data. Please note that the VIRAT dataset consists only of RGB videos and no depth maps. Furthermore, depth estimation in an outdoor setting is useful as it is an appearance invariant feature (similar to optical flow) which are known to be helpful in activity recognition. ADR shows that using simulation, monocular outdoor depth estimation is possible without installing costly depth sensors.https://www.overleaf.com/project/5c7c3e22ac6a080f4fd3ff46

\begin{figure*}[t!]
\begin{center}
\begin{tabular}{cc}
 \includegraphics[width=0.45\linewidth, height=0.20\linewidth]{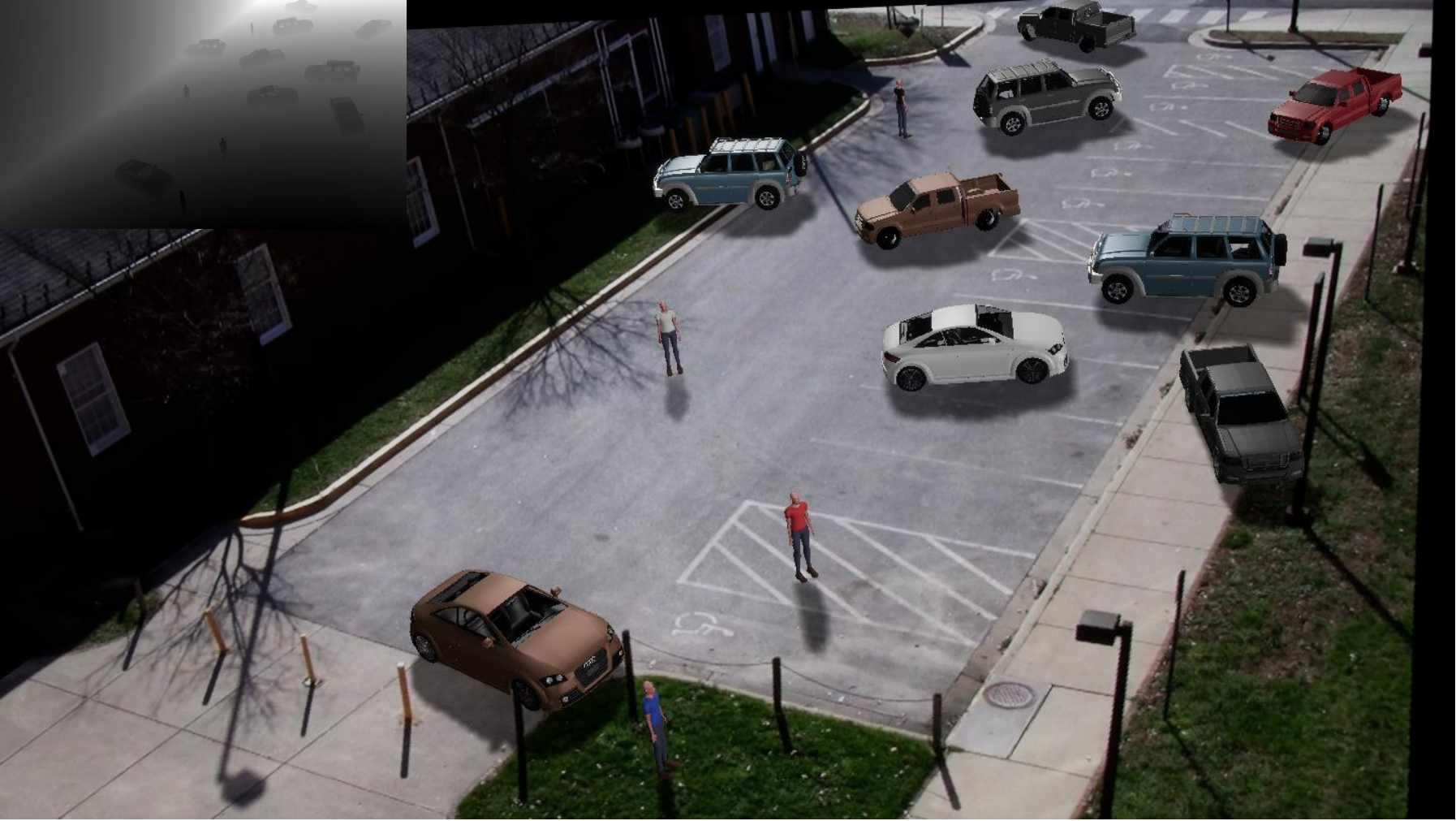} &
 \includegraphics[width=0.45\linewidth, height=0.20\linewidth]{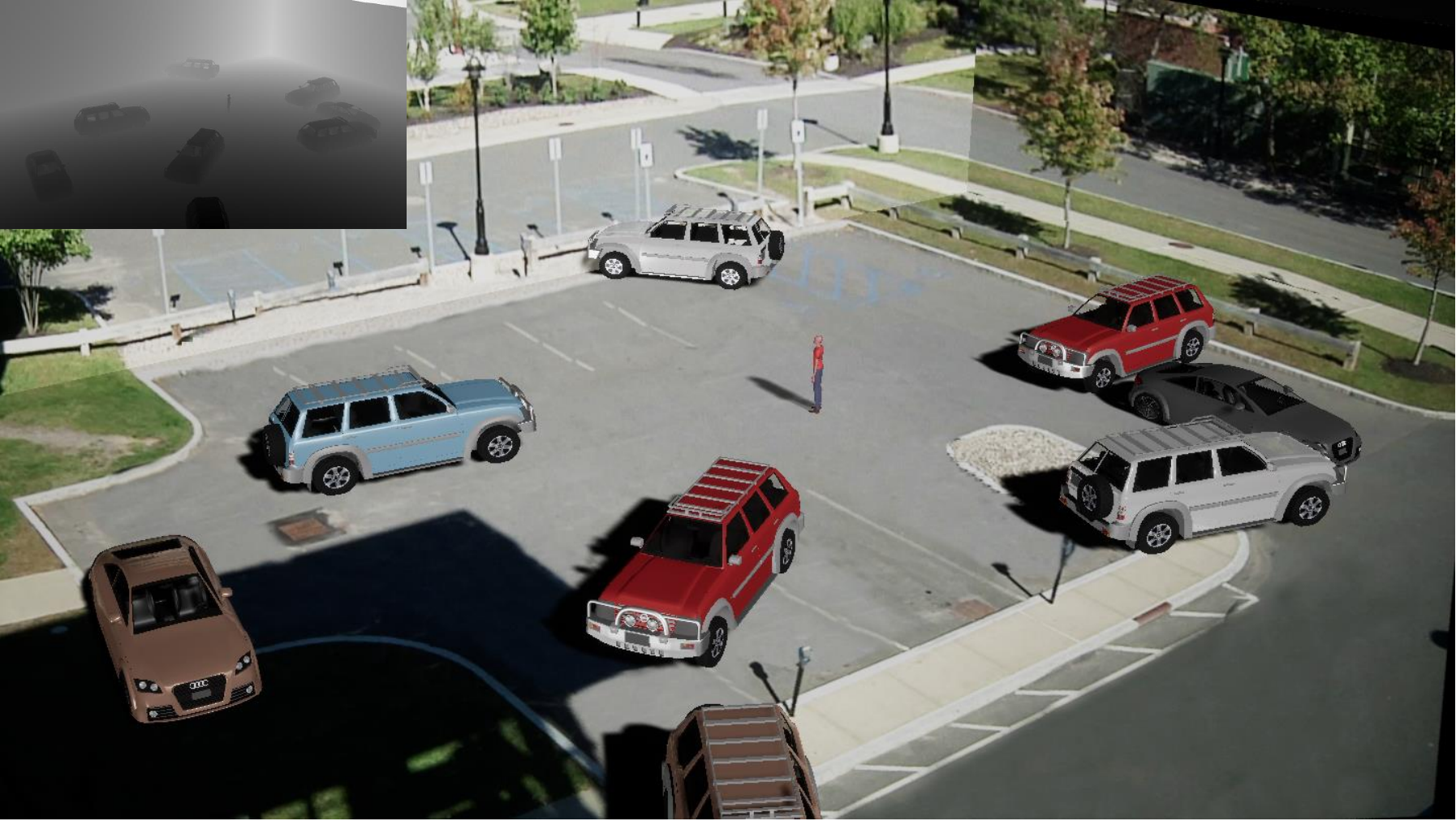}
 \end{tabular}
\end{center}
   \caption{Source data for VIRAT scenes rendered using VADRA along with the ground truth depth map.}
\label{fig:ADR_depth_samples}
\end{figure*}

\begin{figure*}[t]
\begin{center}
\begin{tabular}{cc}
 \includegraphics[width=0.49\linewidth]{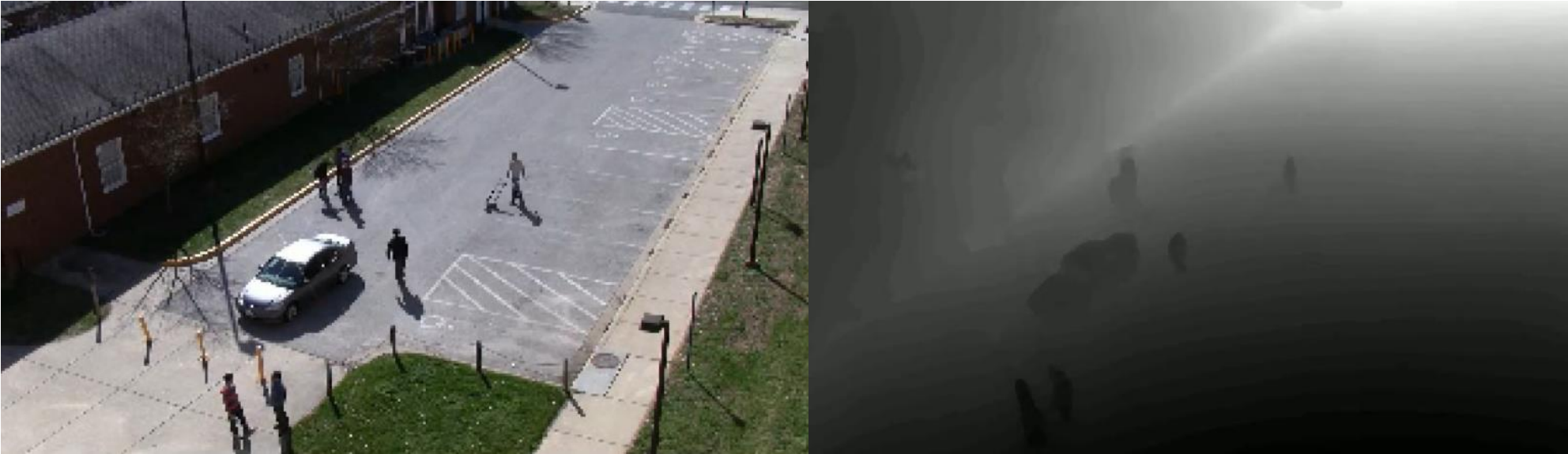} 
 \includegraphics[width=0.49\linewidth]{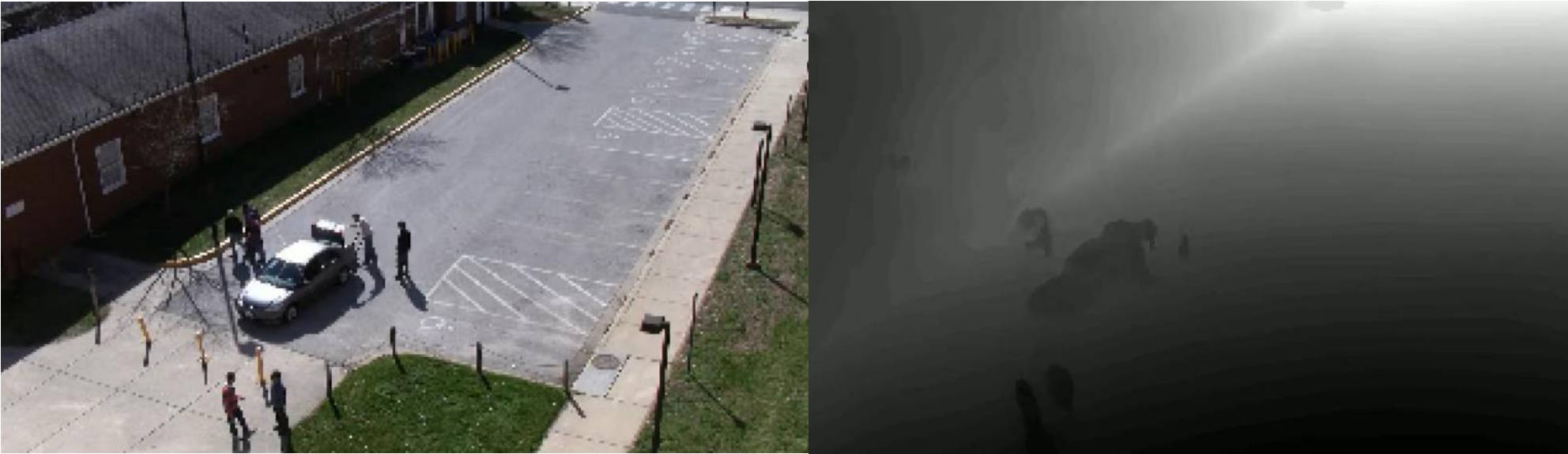} \\
 \includegraphics[width=0.49\linewidth]{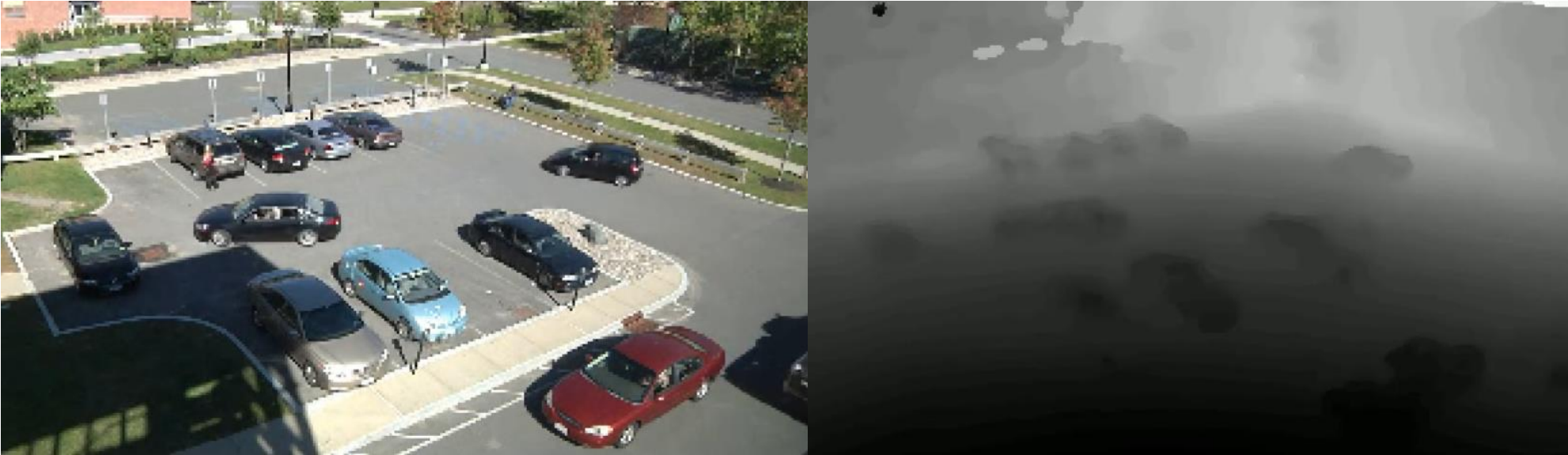}
 \includegraphics[width=0.49\linewidth]{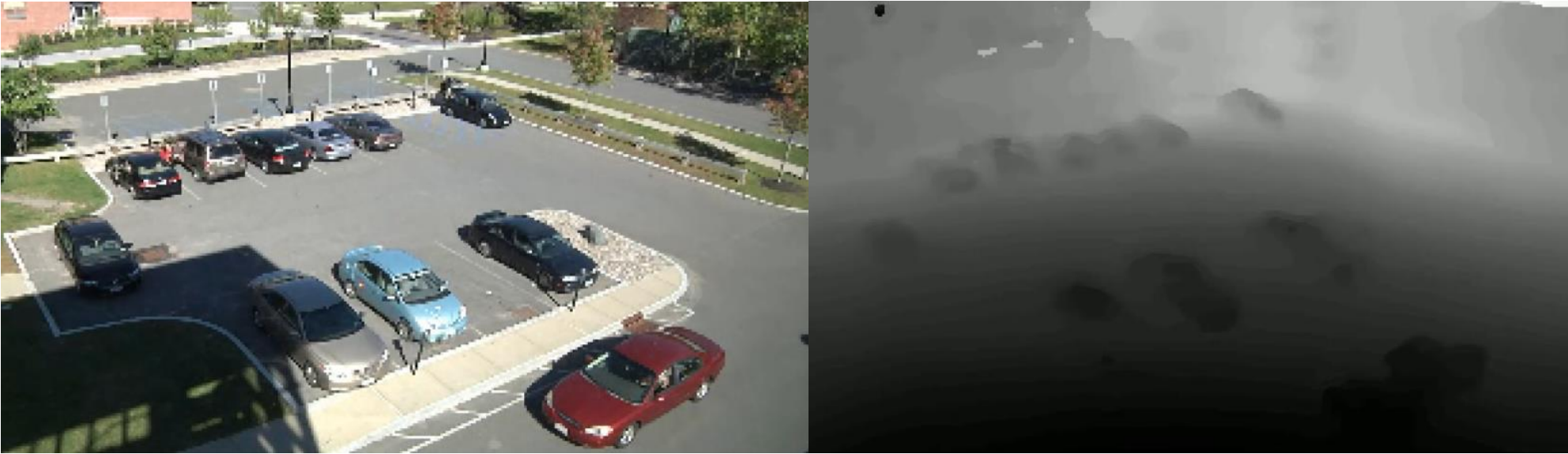}
\end{tabular}
\end{center}
\vspace{-2mm}
   \caption{Depth estimation output from an FCN trained on data generated by ADR for VIRAT} 
\label{fig:depth_results}
\end{figure*}

 The input space $\mathcal{X}$ consists of images with resolution $1920 \times 1080$ and the output space $\mathcal{Y}$ is the space of depth images of resolution $454 \times 256$, where each pixel corresponds to the density quantized in 80 bins. Here $\theta \in \Theta$ (similar to object detection) is a list of object attributes in the image. These attributes specify the location and type of the object in the image. Fig. \ref{fig:ADR_depth_samples} shows labeled samples from source domain.

\textbf{ADR Setup:}
The setup is similar to object detection but we do not use distractors for this task and only choose between person or car object. The reward is again computed per cell and is the negative of the average cross entropy loss using the learner $h$'s predictions.

The learner $h$ is implemented as an FCN \cite{long2015fully} with a ResNet101 \cite{he2016deep} as the backbone architecture. $h$ classifies each pixel into one of the bins at a coarse resolution of $454 \times 256$. The batch size is set to 2 along with SGD optimizer with a linearly decaying learning rate. We train a separate model for each scene in VIRAT.

\textbf{Results}: Figure \ref{fig:depth_results} shows qualitative results for monocular depth estimation. $h$ very quickly learns the depth profile of the background as it is almost constant for all the source domain. $\pi_\omega$ attempts to make it harder for $h$ to predict the depth profile of the foreground object by minimizing its pixels in the image. This results into increasing the spawn probability of cells away from the camera. As a result, our task model trained using ADR even accurately predicts the depth profile of small foreground objects like people.

{\small
\bibliographystyle{ieee}
\bibliography{supplementary_references}
}

%% file: sections/introduction.tex


\begin{figure}[t!]
\centering
\begin{center}
\resizebox{3.5in}{!}{%
\begin{tabular}{c}
\includegraphics[width=0.95\linewidth]{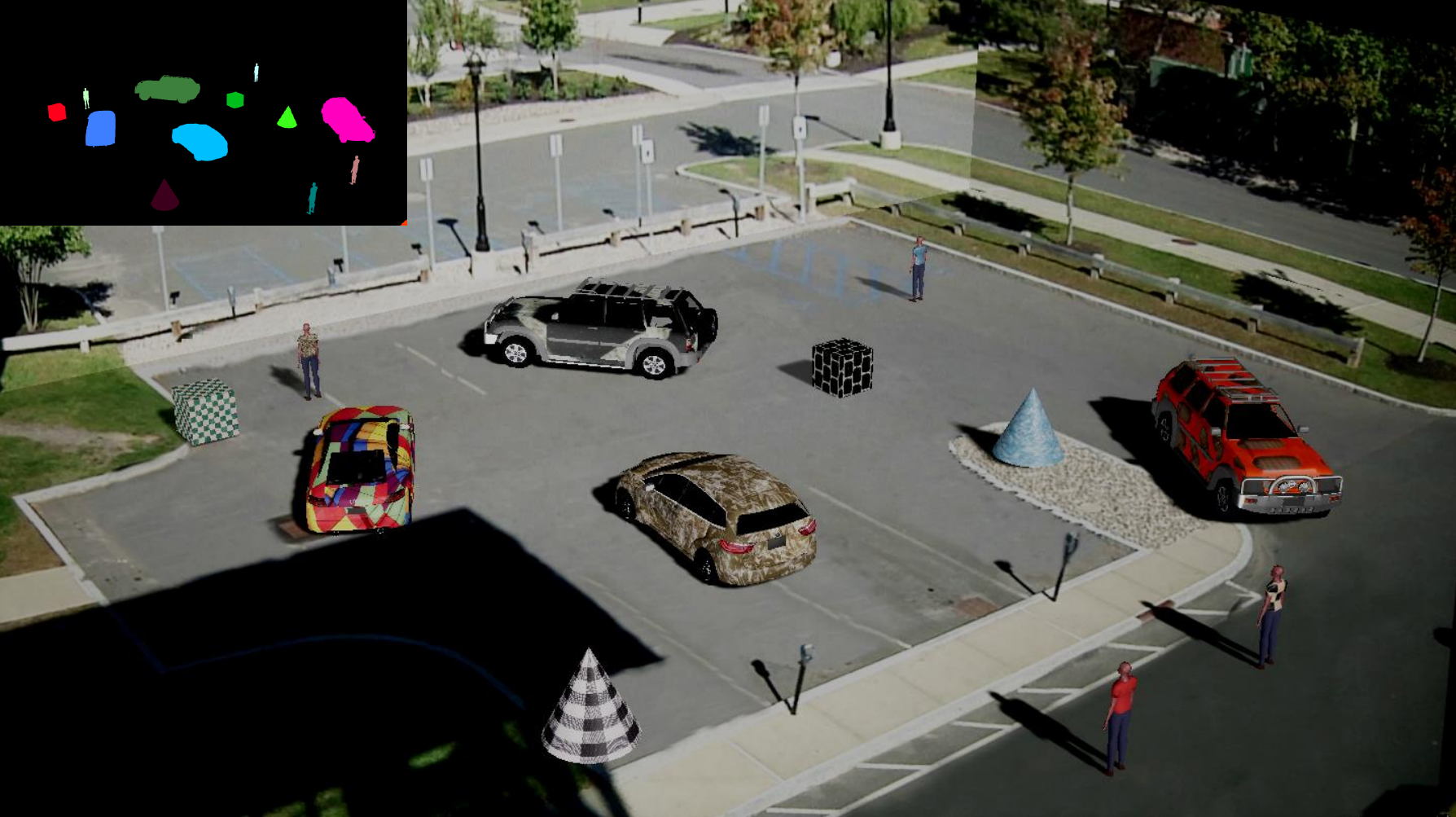}
\label{fig:intro_det}\\
(a) Domain Randomization (DR)\\\\
\includegraphics[width=0.95\linewidth]{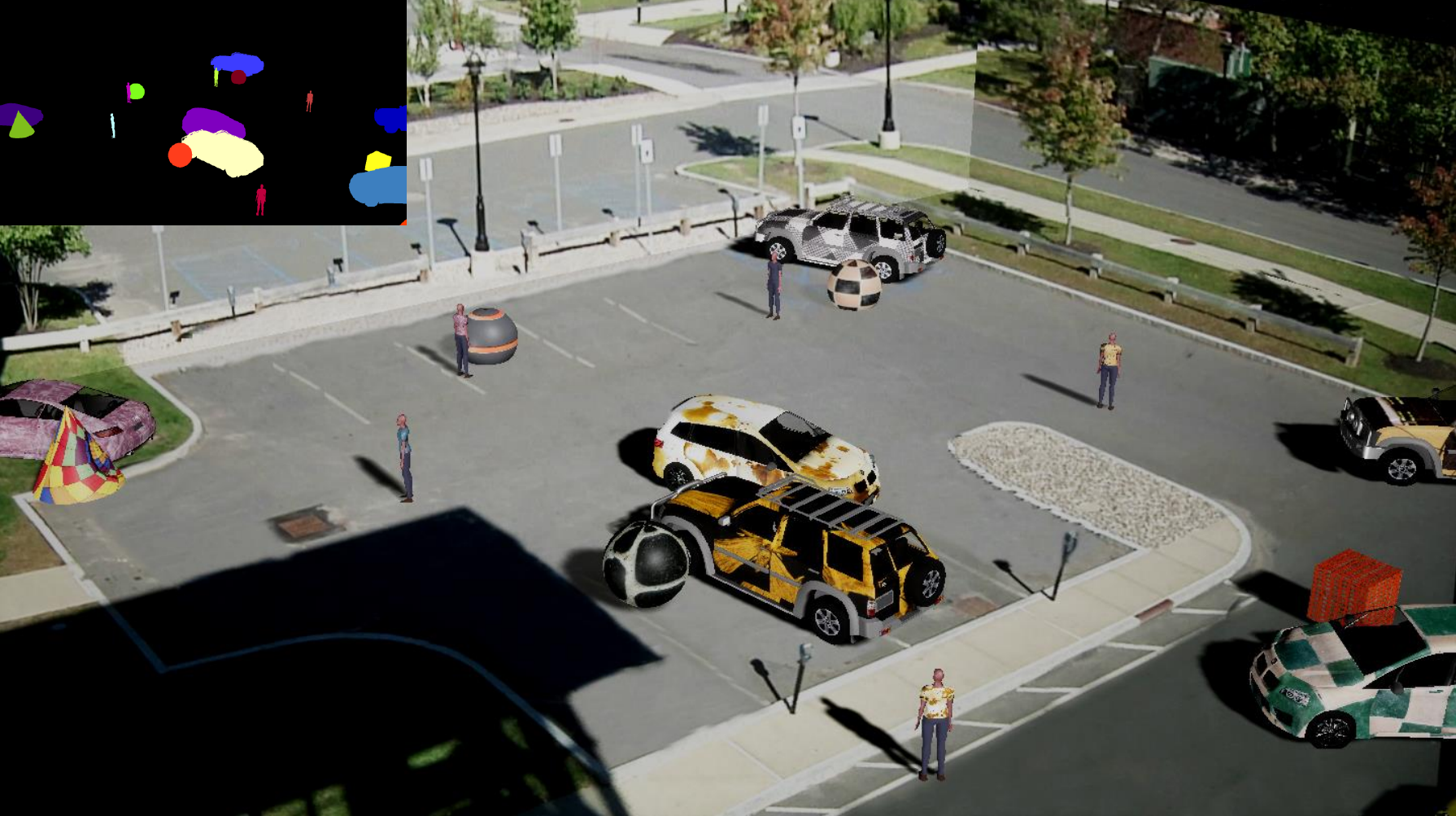}
\label{fig:intro_depth}\\
(b) Adversarial Domain Randomization (ADR)
 \end{tabular}
}
\end{center}
  \caption{We compare  synthetic images generated by DR and ADR along with their instance segmentation maps for the task of object detection. ADR learns to generate harder samples containing occluded and truncated objects. This allows efficient learning with few samples. }
\vspace{-6mm}
\label{fig:ADR_intro}
\end{figure}

    
A large amount of labeled data is required to train deep neural networks. However, a manual annotation process on such a scale can be expensive and time-consuming, especially for complex vision tasks like semantic segmentation where labels are difficult to specify. It is, thus, natural to explore simulation for generating synthetic data as a cheaper and quicker alternative to manual annotation. However, while use of simulation makes data annotation easier, it results in the problem of \textit{domain gap}. This problem arises when the training data distribution - \textit{source domain} differs from the test data distribution - \textit{target domain}. In this case, learners trained on synthetic data would suffer from high generalization error. There are two popular solutions to reducing domain gap, (1) the well studied paradigm of domain adaptation (DA) \cite{wang2018deep} and (2) the recently proposed domain randomization (DR) \cite{tobin2017domain}. Both the solutions focus on extracting \textit{domain invariant features} from labeled source domain to bridge the gap. DA assumes access to unlabeled target data and achieves domain invariant feature extraction by explicitly minimizing feature discrepancy between source and target data. On the other hand, in the absence of unlabeled target data, DR bridges the gap by adding enough variations (\eg random textures onto objects for detection) to source domain. These variations force the learner to focus on essential domain invariant features.


A natural question therefore arises: how does DR compensate for the lack of information on target domain? The answer lies in the design of the simulator used by DR for randomization. The simulator is encoded with target domain knowledge (\eg priors on shape of objects) beforehand. This acts as a sanity check between the labels of source domain and target domain (\eg cars should not be labeled as trucks). Another key design decision involves which simulation parameters are supposed to be randomized. This plays an important role in deciding the domain invariant features which help the learner in generalizing to the target domain. Clearly, these domain invariant features are task dependent, as they contain critical information to solve the task. \eg in car detection, the shape of a car is a critical feature and has to be preserved in source domain, it is therefore acceptable to randomize the car's appearance by adding textures in DR. However, when the task changes from car detection to a red car detection, car's appearance becomes a critical feature and can no longer be randomized in DR. We believe it is important to understand such characteristics of the DR algorithm. Therefore, as a step in this direction, in section \ref{section:thoery_dr}, we provide theoretical analysis of DR. Concretely our analysis answers the following questions about DR: 1) how does it make the learner focus on domain invariant features? 2) what is its generalization error bound? 3) in presence of unlabeled target data, can we use it with DA? 4) what are its limitations?


In our analysis, we identify a key limitation: DR requires a lot of data to be effective. This limitation mainly arises from DR's simple strategy of using a uniform distribution for simulation parameter randomization. As a result, DR often generates samples which the learner has already seen and is good at. This wastes valuable computational resources during data generation. We can view this as an exploration problem in the space of simulation parameters. A uniform sampling strategy in this space clearly does not guarantee sufficient exploration in image space. We address this limitation in section \ref{section:adr} by proposing Adversarial Domain Randomization (ADR), a data efficient variant of DR. ADR generates adversarial samples with respect to the learner during training. We implement ADR as a policy whose action space is the quantized simulation parameter space. At each iteration, the policy is updated using policy gradients to maximize learner's loss on generated data whereas the learner is updated to minimize this loss. As a result, we observe ADR frequently generates samples like truncated and occluded objects for object detection and confusing classes for image classification. Figure~\ref{fig:ADR_intro} shows a comparison of sample images generated by DR and ADR for the task of object detection.

Finally, we go back to the question posed previously, can we use DR with DA when unlabeled target data is available? Our reinforcement learning framework for ADR easily extends to incorporate DA's feature discrepancy minimization in form of reward for the policy. We thus incentivize the policy to generate data which is consistent with target data. As a result, we no longer need to manually encode domain knowledge in the design of the simulator. In summary, the contributions of our work are as follows: 

\begin{itemize}[leftmargin=*]
\setlength\itemsep{-1mm}
    \item \textbf{Theoretical Analysis for Domain Randomization}: We present a theoretical perspective on effectiveness of DR. We also provide a bound for its generalization error and analyze its limitations.
    \item \textbf{Adversarial Domain Randomization}: As a solution to DR's limitations, we propose a more data efficient variant of DR, namely ADR.
    \item \textbf{Evaluations on Real Datasets for Diverse Tasks}: We benchmark our approach on image classification and object detection on real-world datasets like Syn2Real \cite{peng2018syn2real}, VIRAT \cite{sangmin2011virat}.
\end{itemize}

%% file: sections/related_work.tex
Our work is broadly related to approaches using a simulator as a source of supervised data and solutions for the reduction of domain gap.

\vspace{1mm}
\noindent\textbf{Synthetic Data for Training} Recently with the advent of rich 3D model repositories like ShapeNet and the related ModelNet \cite{chang2015shapenet}, Google 3D warehouse \cite{arlinghaus2007google}, ObjectNet3D \cite{xiang2016objectnet3d}, IKEA3D \cite{lim2013parsing}, PASCAL3D+ \cite{xiang2014beyond} and increase in accessibility of rendering engines like Blender3D, Unreal Engine 4 and Unity3D, we have seen a rapid increase in using synthetic data for performing visual tasks like object classification \cite{peng2018syn2real}, object detection \cite{peng2018syn2real, tremblay2018training, hattori2018pedestrian}, pose estimation \cite{Su_2015_ICCV, kundu20183d, sundermeyer2018implicit}, semantic segmentation \cite{varol2017learning, saleh2018effective} and visual question answering \cite{johnson2017clevr}. Often the source of such synthetic data is a simulator, and use of simulators for training control policies is already a popular approach in robotics \cite{brockman2016openai, tassa2018deepmind}. SYNTHIA \cite{ros2016synthia}, GTA5 \cite{richter2016playing}, VIPER \cite{richter2017playing}, CLEVR \cite{johnson2017clevr}, AirSim \cite{airsim2017fsr}, CARLA \cite{dosovitskiy2017carla}  are some of the popular simulators in computer vision.


\vspace{1mm}
\noindent\textbf{Domain Adaptation}: 
Given source domain and target domain, methods like \cite{bousmalis2016domain, chen2018road, hoffman2017cycada, hoffman2016fcns, zhang2017curriculum, murez2017image, tzeng2017adversarial} aim to reduce the gap between the feature distributions of the two domains. \cite{bousmalis2016domain, ganin2016domain, ganin2014unsupervised, tzeng2017adversarial, hoffman2016fcns} did this in an adversarial fashion using a discriminator for domain classification whereas \cite{tzeng2014deep, long2015learning} minimized a defined distance metric between the domains. Another approach is to match statistics on the batch, class or instance level \cite{hoffman2016fcns, chen2018domain} for both the domains. Although these approaches outperform simply training on source domain, they all rely on having access to target data albeit unlabelled.

\vspace{1mm}
\noindent\textbf{Domain Randomization}: 
These methods \cite{sadeghi2016cad2rl, dosovitskiy2016learning, tobin2017domain, james2017transferring, tremblay2018training, pinto2017asymmetric, peng2018sim, tan2018sim, sundermeyer2018implicit, khirodkar2018domain} do not use any information about the target domain during training and only rely on a simulator capable of generating varied data. The goal is to close the domain gap by generating synthetic data with sufficient variation that the network views real data as just another variation. The underlying assumption here is that simulator encodes the domain knowledge about the target domain which is often specified manually \cite{prakash2018structured}.

%% file: sections/theoretical_analysis.tex
We analyze DR using generalization error bounds from multiple source domain adaptation \cite{ben2010theory}. The key insight is to view DR as a learning problem from multiple source domains where each source domain represents a particular subset of the data space. For example, consider the space of all car images. The images containing a car of a particular make form a subset of that space which we refer to as a single source domain. If one were to generate random images across different car makes, as we do in DR, this can be interpreted as combining data from multiple source domains. 

In this section, we first introduce the preliminaries in \ref{subsection:prelim} followed by a formal definition of DR algorithm in \ref{subsection:DR}. Lastly, we draw parallels between DR and multiple source domain adaptation in \ref{subsection:dr_equal_source} where we also show that the generalization error bound for DR is better than the bound for data generation without randomization.

\subsection{Preliminaries}\label{subsection:prelim}

The notation introduced below is based on the theoretical model for DA using multiple sources for binary classification \cite{ben2010theory}. The analysis here is limited to binary classification for simplicity but can be extended to other tasks as long as the triangle inequality holds \cite{ben2007analysis}. 

A \textit{domain} is defined as a tuple $\left \langle \mathcal{D}, f \right \rangle$ where: (1) $\mathcal{D}$ is a distribution over input space $\mathcal{X}$ and (2) $f: \mathcal{X} \mapsto \mathcal{Y}$ is a labeling function, $\mathcal{Y}$ being the output space which is $[0,1]$ for binary classification. $N$ source domains are denoted as $\{ \left \langle \mathcal{D}_i, f_i \right \rangle \}|_{i=1}^{N}$ and the target domain is denoted as $\left \langle \mathcal{D}_T, f_T \right \rangle$. A \textit{hypothesis} is a binary classification function $h: \mathcal{X} \mapsto \{0,1\}$. The \textit{error} (sometimes called \textit{risk}) of a hypothesis $h$ w.r.t a labeling function $f$ under distribution $\mathcal{D}$ is defined as $\epsilon(h, f, \mathcal{D}) := \mathbb{E}_{x\sim \mathcal{D}}\big[|h(x)-f(x)|\big]$. We denote the error of hypothesis $h$ on target domain as $\epsilon_T(h) = \epsilon(h, f_T, \mathcal{D}_T)$ and on $i^{\text{th}}$ source domain as $\epsilon_i(h) = \epsilon(h, f_i, \mathcal{D}_i)$. As common notation in computational learning theory, we use $\epsilon_T(h)$ and $\hat{\epsilon}_T(h)$ to denote the true error and empirical error on the target domain. Similarly, $\epsilon_i(h)$ and $\hat{\epsilon}_i(h)$ are defined for $i^{\text{th}}$ source domain.

\vspace{1mm}
\noindent
\textbf{Multiple Source Domain Problem.} Our goal is to learn a hypothesis $h$ from hypothesis class $\mathcal{H}$ which minimizes $\epsilon_T(h)$ on the target domain $\left \langle \mathcal{D}_T, f_T \right \rangle$ by only using labeled samples from $N$ source domains $\{ \left \langle \mathcal{D}_i, f_i \right \rangle \}|_{i=1}^{N}$. 

\vspace{1mm}
\noindent
\textbf{$\boldsymbol\alpha$-Source Domain.} We combine $N$ source domains into a single source domain denoted as $\alpha$-source domain where $\alpha$ helps us control the contribution of each source domain during training. We denote by $\Delta$ the simplex of $\mathbb{R}^N$, $\Delta = \{\alpha: \alpha_i \geq 0 \wedge \sum_{i=1}^{N}\alpha_i = 1 \}$. Any $\alpha \in \Delta$ forms an $\alpha$-source domain. The error of a hypothesis $h$ on this source domain ($\alpha$-error) is denoted as $\epsilon_\alpha(h) = \sum_{i=1}^{N} \alpha_i \epsilon_i(h)$. The $\alpha$ input distribution is denoted as $\mathcal{D}_\alpha = \sum_{i=1}^{N}\alpha_i \mathcal{D}_i$.

The multiple source domain problem can now be reduced to training on labeled samples from $\alpha$-source domain for various values of $\alpha \in \Delta$. We use $\mathcal{D}_\alpha$ to sample inputs from $\mathcal{X}$, which are then labeled using all the labeling functions $\{ \left \langle f_i \right \rangle \}|_{i=1}^{N}$. We learn a hypothesis $h$ to minimize the empirical $\alpha$-error, $\hat{\epsilon}_\alpha(h) = \sum_{i=1}^{N} \alpha_i \hat{\epsilon}_i(h)$. 

\vspace{1mm}
\noindent
\textbf{Generalization Error Bound.} Let $\hat{h}_\alpha = \argmin_h \hat{\epsilon}_\alpha(h)$ and $h^*_\alpha = \argmin_h \epsilon_\alpha(h)$ \ie $\hat{h}_\alpha$ and $h^*_\alpha$ minimize the empirical and true $\alpha$-error respectively. We are interested in bounding the generalization error $\epsilon_T(\hat{h}_\alpha)$ for empirically optimal hypothesis $\hat{h}_\alpha$. However, using Hoeffding's inequality \cite{serfling1974probability}, it can be shown that minimum empirical error $\hat{\epsilon}_\alpha(\hat{h}_\alpha)$ converges uniformly to the minimum true error $\epsilon_\alpha(h^*_\alpha)$ $\ie$ without loss of generality $\hat{h}_\alpha$ converges to $h^*_\alpha$ given large number of samples. To simplify our analysis we instead bound generalization error $\epsilon_T(h^*_\alpha)$ for true optimal hypothesis $h^*_\alpha$ (the bound for $\hat{h}_\alpha$ is provided in supplementary material).

Following the proof for Th. 5 (\textit{A bound using combined divergence}) in \cite{ben2010theory}, we provide a bound for generalization error $\epsilon_T(h^*_\alpha)$ below.

\begin{theorem}
\label{theorem:generalization_bound} {(Based on Th.5 \cite{ben2010theory})}
Consider the optimal hypothesis on target domain $h^*_T = \argmin_h \epsilon_T(h)$ and on $\alpha$-source domain $h^*_\alpha = \argmin_h \epsilon_\alpha(h)$. If $\gamma_\alpha = \min_h \{ \epsilon_T(h) + \epsilon_\alpha(h) \}$, then
\begin{align*}
\epsilon_{T}(h^*_\alpha) & \leq \epsilon_T(h^*_T) + 2\gamma_\alpha + d_{\mathcal{H}\Delta\mathcal{H}} (\mathcal{D}_{\alpha}, \mathcal{D}_T)
\end{align*}
\end{theorem}

\noindent
\textbf{Remarks}: Proof in supplementary material. $\epsilon_T(h^*_T)$ is the minimum true error possible for the target domain (clearly, $\epsilon_T(h^*_T) \leq \epsilon_T(h^*_\alpha)$). $\gamma_\alpha$ represents the minimum error using both the target domain and $\alpha$-source domain jointly. Intuitively, it represents the agreement between all the labeling functions involved (target domain and all source domains) \ie $\gamma_\alpha$ would be large, if these labeling functions label an input differently. Lastly, $d_{\mathcal{H}\Delta\mathcal{H}} (\mathcal{D}_{\alpha}, \mathcal{D}_T)$ is the $\mathcal{H}\Delta\mathcal{H}$ divergence between input distributions $\mathcal{D}_{\alpha}$ and $\mathcal{D}_T$. In summary, generalization on target domain $\epsilon_T(h^*_\alpha)$ depends on: (1) difficulty of task on target domain $\epsilon_T(h^*_T)$; (2) labeling consistency between target domain and source domains $\gamma_\alpha$; (3) similarity of input distribution between target and source domains $d_{\mathcal{H}\Delta\mathcal{H}} (\mathcal{D}_{\alpha}, \mathcal{D}_T)$.

\subsection{Domain Randomization}\label{subsection:DR}

DR addresses the multiple source domain problem by modeling various source domains using a simulator with randomization. The simulator is a generative module which produces labeled data $(x,y)$. In practice, DR uses an accurate simulator which internally encodes the knowledge about the target domain as a target labeling function $f_T(x)$, such that $y = f_T(x)$.  Concretely, let $\Theta$ be the rendering parameter space and $\mathcal{D}_\Theta$ be a probability distribution over $\Theta$. We denote the simulator as a function $g: \Theta \mapsto \mathcal{X} \times \mathcal{Y}$ such that $g(\theta) = \{x,y\}$ where $\theta \sim \mathcal{D}_\Theta$. Simply put, a simulator $g$ takes a set of parameters $\theta$ and generates an image $x$ and its label $y$. In general, the DR algorithm generates data by randomly (uniformly) sampling $\theta$ from $\Theta$ \ie $\mathcal{D}_\Theta$ is set to $\mathcal{U}_\Theta$, an uniform distribution over $\Theta$ (refer Alg. \ref{dr_algo}). The algorithm outputs a hypothesis $\hat{h}$ which empirically minimizes the loss $\ell(\hat{h}(x_i), y_i)$ over $M$ data samples. Note, in our analysis, we set $\ell(\hat{h}(x), y)$ to be $|\hat{h}(x) - y|$. \textsc{learner-update} is the parameter update of $\hat{h}$ using loss $\ell(\hat{h}(x_i), y_i)$. 

\begin{algorithm}
\caption{Domain Randomization}\label{dr_algo}
\hspace*{\algorithmicindent} \textbf{Input:} $g, M$ \\
\hspace*{\algorithmicindent} \textbf{Output:} $\hat{h}$ 
\begin{algorithmic}[1]
\For{$i \in \{1,2,\ldots,  M\}$}
    \State $\theta \sim \mathcal{U}_\Theta$
    \State $\{x,y\} = g(\theta)$
    \State $\hat{h} = \textsc{learner-update}(\hat{h}, x, y)$
\EndFor
\end{algorithmic}
\end{algorithm}

\noindent
The objective function optimized by these steps can be written as follows:
\begin{gather*}
\min\limits_{h \in \mathcal{H}} \displaystyle \mathop{\mathbb{E}}_{\theta \sim \mathcal{U}_{\Theta}}\Big[\ell \Big(h\big(g(\theta)_x \big), g(\theta)_y \Big)\Big] \tag{1}\label{eq:dr-loss}
\end{gather*}
\subsection{DR as $\boldsymbol\bar{\alpha}$-Source Domain}\label{subsection:dr_equal_source}
We interpret data generated using DR as labeled data from an $\alpha$-source domain, specifically $\alpha = [\frac{1}{N}, \ldots,\frac{1}{N}]$ (referred as $\bar{\alpha}$ hereafter). This captures equal contribution by each sample during training according to Alg \ref{dr_algo}. Using Th. \ref{theorem:generalization_bound}, we can bound the generalization error for $\bar{\alpha}$-source domain (DR) by $\epsilon_T(h^*_T) + 2 \bar{\gamma} + d_{\mathcal{H}\Delta\mathcal{H}} (\mathcal{D}_{\bar{\alpha}}, \mathcal{D}_T)$ where $\bar{\gamma} = \min_h \{ \epsilon_T(h) + \frac{1}{N} \sum_{i=1}^{N} \epsilon_i(h)\}$.

We now compare DR with data generation without randomization or variations. The later is same as choosing only one source domain for training, denoted as $\alpha_i$-source domain where $\alpha_i$ is a one-hot $N$-vector indicating domain $i \in \{1,\ldots,N\}$. The generalization error bound for $\alpha_i$-source domain would be $\epsilon_T(h^*_T) + 2 \gamma_i + d_{\mathcal{H}\Delta\mathcal{H}} (\mathcal{D}_{i}, \mathcal{D}_T)$ where $\gamma_i = \min_h \{\epsilon_T(h) + \epsilon_i(h) \}$.

\begin{figure}
\begin{center}
\includegraphics[width=0.6\linewidth]{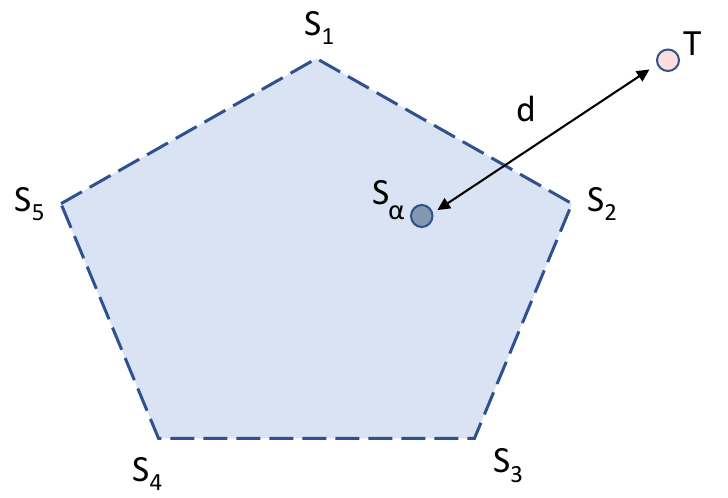}
\end{center}
   \caption{A visualization of the domains using simplex $\Delta (N = 5$). The corners of $\Delta$ are the source domains $S_1, \ldots, S_5$ and the point $T$ is the target domain. Any interior point $S_\alpha \in \Delta$ is the upper bound for generalization error $\epsilon_T(h^*_\alpha)$ and the point T is $\epsilon_T(h^*_T)$. The distance between $S_\alpha$ and $T$ is $d = 2 \gamma_\alpha + d_{\mathcal{H}\Delta\mathcal{H}} (\mathcal{D}_{\alpha}, \mathcal{D}_T)$. The data generated using DR is the centre of $\Delta$.} 
\label{fig:distance}
\end{figure}


Refer to Fig. \ref{fig:distance} for a visualization of generalization error of $S_\alpha$ as a distance measure from target domain, we define $d = 2 \gamma_\alpha + d_{\mathcal{H}\Delta\mathcal{H}} (\mathcal{D}_{\alpha}, \mathcal{D}_T)$ as the upper bound for $|\epsilon_T(h^*_\alpha) - \epsilon_T(h^*_T)|$. We wish to find an optimal $\alpha$ which minimizes this distance measure \ie the point in $\Delta$ closest to the target domain ($\alpha_2$ in Fig. \ref{fig:distance}). However, when no unlabeled target data is available it is best to choose the centre of $\Delta$ ($\alpha = \bar{\alpha})$ as our source domain. We prove this in lemma \ref{lemma:comparison}, which states that the distance of target domain from the centre of $\Delta$ is less than the average distance from the corners of $\Delta$.

\begin{lemma} For $i \in \{1,\ldots, N\}$, let $\alpha_i = [0,.. \underset{i^{\text{th}}}{1}.. 0] $, $\gamma_i = \min_h \{\epsilon_T(h) + \epsilon_i(h) \}$  and $\bar{\alpha} = [\frac{1}{N}, \frac{1}{N}..., \frac{1}{N}] $, $\bar{\gamma} = \min_h \{ \epsilon_T(h) + \frac{1}{N} \sum_{i=1}^{N} \epsilon_i(h) \}$, then
\label{lemma:comparison}

\vspace{-5mm}
\begin{align*}
\frac{1}{N} \sum_{i=1}^{N} \Big( 2\gamma_i + d_{\mathcal{H}\Delta\mathcal{H}} (\mathcal{D}_i, \mathcal{D}_T)\Big) \geq 2\bar{\gamma} + d_{\mathcal{H}\Delta\mathcal{H}} (\mathcal{D}_{\bar{\alpha}}, \mathcal{D}_T)
\end{align*}
\end{lemma}

\noindent
\textbf{Remarks:} Proof provided in supplementary material follows from the convexity of distance measure $2\gamma_i + d_{\mathcal{H}\Delta\mathcal{H}} (\mathcal{D}_i, \mathcal{D}_T)$ with application of Jensen's inequality \cite{liao2018sharpening}. Using this lemma, the corollary \ref{corollary:DR_single} states that in the absence of unlabeled target data, in expectation DR (centre of $\Delta$) is superior to data generation without randomization (any other point in $\Delta$).

\begin{corollary}
The generalization error bound for $\bar{\alpha}$-source domain (DR) is smaller than the
expected generalization error bound of a single source domain (expectation over a uniform choice of source domain). 
\label{corollary:DR_single}
\end{corollary}



%% file: sections/method.tex

We modify DR's objective ({eq.\ref{eq:dr-loss}}) by making a pessimistic (adversarial) assumption about $\mathcal{D}_\Theta$ instead of assuming it to be stationary and uniform. By making this adversarial assumption, we force the learned hypothesis to be robust to adversarial variations occurring in the target domain. This type of worst case modeling is especially desirable when annotated target data is not available for rare scenarios.

The resulting min-max objective function is as follows:
\begin{gather*}
\min\limits_{h \in \mathcal{H}} \max\limits_{\mathcal{D}_\Theta} \displaystyle \mathop{\mathbb{E}}_{\theta \sim \mathcal{D}_{\Theta}}\Big[\ell \Big(h\big(g(\theta)_x \big), g(\theta)_y \Big)\Big] \tag{2}
\end{gather*}

\subsection{ADR via Policy Gradient Optimization}

\begin{figure}
 \begin{center}
  \includegraphics[width=\linewidth]{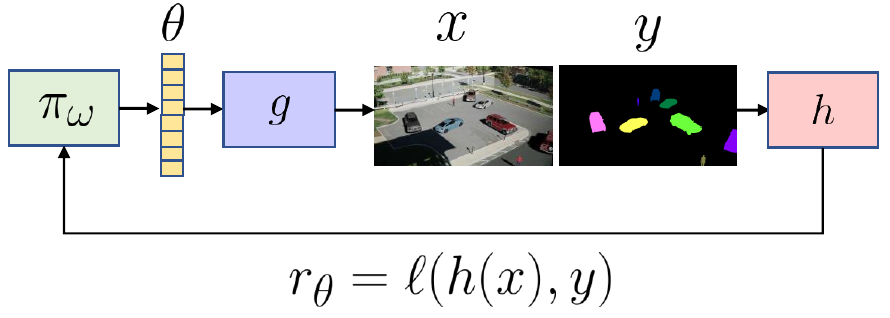}
 \end{center}
   \caption{$\pi_\omega$ is the policy with parameter $\omega$, $g$ is the simulator which takes $\theta$ as input and generates a labeled data sample $x,y$. The learner $h$ is trained to minimize $\ell(h(x), y)$ on the data sample. $\pi_\omega$ is rewarded to maximize the learner's loss.}
 \label{fig:method_ADR}
 \end{figure}

This adversarial objective function is a zero-sum two player game between \textsc{simulator} ($g$) and \textsc{learner} ($h$). The \textsc{simulator} selects a distribution $\mathcal{D}_{\Theta}$ for data generation and the \textsc{learner} chooses $h \in \mathcal{H}$ which minimizes loss on the data generated from $\mathcal{D}_{\Theta}$. The Nash equilibrium of this game corresponds to the optima of the min-max objective, which we find by using reinforcement learning with policy gradients \cite{sutton2000policy}. The \textsc{simulator}'s action $\mathcal{D}_\Theta$ is modeled as the result of following the policy $\pi_\omega$ with parameters $\omega$. $g$ samples $\theta$ according to $\pi_\omega$, which is then converted into labeled data $(x,y)$. The \textsc{learner}'s action $h$ is optimized to minimize loss $\ell(h(x),y)$. The reward $r_\theta$ for policy $\pi_\omega$ is set to this loss. Specifically, we maximize the objective $J(\omega)$ by incrementally updating $\pi_\omega$, where
\begin{align*}
    J(\omega) & =  \displaystyle \mathop{\mathbb{E}}_{\theta \sim \pi_\omega}[r(\theta)] & \text{where} \quad r(\theta) & = \ell\Big(h(g(\theta)_x), g(\theta)_y \Big).
\end{align*}
We use REINFORCE \cite{williams1992simple} to obtain gradients for updating $\omega$ using an unbiased empirical estimate of $\nabla_\omega J(\omega)$ 
\begin{align*}
    \hat{J}(\omega) & = \frac{1}{M} \sum_{i=1}^{m} \nabla_\omega \log(\pi_\omega(\theta)) [r(\theta)- b]  \tag{3}
\end{align*}
where $b$ is a baseline computed using previous rewards and $M$ is the data size.
Both \textsc{simulator} ($\pi_\omega$) and \textsc{learner} ($h$) are trained alternately according to Alg. \ref{ADR_algo} shown pictorially in Fig. \ref{fig:method_ADR}.

\begin{algorithm}
\caption{Adversarial Domain Randomization}\label{ADR_algo}
\hspace*{\algorithmicindent} \textbf{Input:} $g$, $M$\\
\hspace*{\algorithmicindent} \textbf{Output:} $\hat{h}$ 
\begin{algorithmic}[1]
\For{$i \in \{1,2, \ldots, M\}$}
    \State \text{$\theta_1 \sim \pi_\omega$}
    \State $\{x_1,y_1\} = g(\theta_1)$
    \State $\hat{h} = \textsc{learner-update}(\hat{h}, x_1,y_1)$\\
    
    \State \text{$\theta_2 \sim \pi_\omega$}
    \State $\{x_2,y_2\} = g(\theta_2)$
    \State $r = \ell(\hat{h}(x_2),y_2)$
    \State $\pi_\omega = \textsc{simulator-update}(\pi_\omega, r, \theta)$ \Comment{Using eq. 3}
\EndFor
\end{algorithmic}
\end{algorithm}

\subsection{DA as ADR with Unlabeled Target Data}

In the original formulation of the ADR problem above, our task was to generate a multi-source data that would be useful for any target domain. An easier variant of the problem exists where we do have access to unlabeled target data. As mentioned before, this falls under the DA paradigm.

To use unlabeled target data, similar to \cite{tzeng2017adversarial} we introduce a domain classifier $D$ which empirically computes $d_{\mathcal{H}\Delta\mathcal{H}} (\mathcal{D}_{\alpha}, \mathcal{D}_T)$. $D$ takes $\phi_h(x)$ as input where $\phi_h$ is a function which extracts feature from input $x$ using $h$. $D$ classifies $\phi_h(x)$ into either from target domain (label 1) or source domain (label 0). The reward function for $\pi_\omega$ is modified to incorporate this distance measure as \begin{align}
    r(\theta) = \ell \Big(h(g(\theta)_x), g(\theta)_y \Big) + w_1 \log{D \Big(\phi_h(g(\theta)_x) \Big)}\nonumber
\end{align}
where $w_1$ is a hyper-parameter. This new reward encourages the policy $\pi_\omega$ to fool $D$, which makes the simulator $g$ generate synthetic data which looks similar to target data. 

However, it is plausible that due to simulator's limitations, we might never be able generate data that looks exactly like target data \ie the simplex $\Delta$ corresponding to $g$ might be very far from point T. In this case, we also modify $h$'s loss as $\ell \Big(h(g(\theta)_x), g(\theta)_y \Big) + w_2 \log{D \Big(\phi_h(g(\theta)_x) \Big)}$ ($w_2$ is a hyper-parameter). As a result, $h$ extracts features $\phi_h(x)$ which are domain invariant. This allows both $g$ and $h$ to minimize distance measure from the target domain.

%% file: sections/experiments.tex

\begin{figure}
\centering
\begin{center}
\begin{tabular}{cc}

\includegraphics[width=0.9\linewidth]{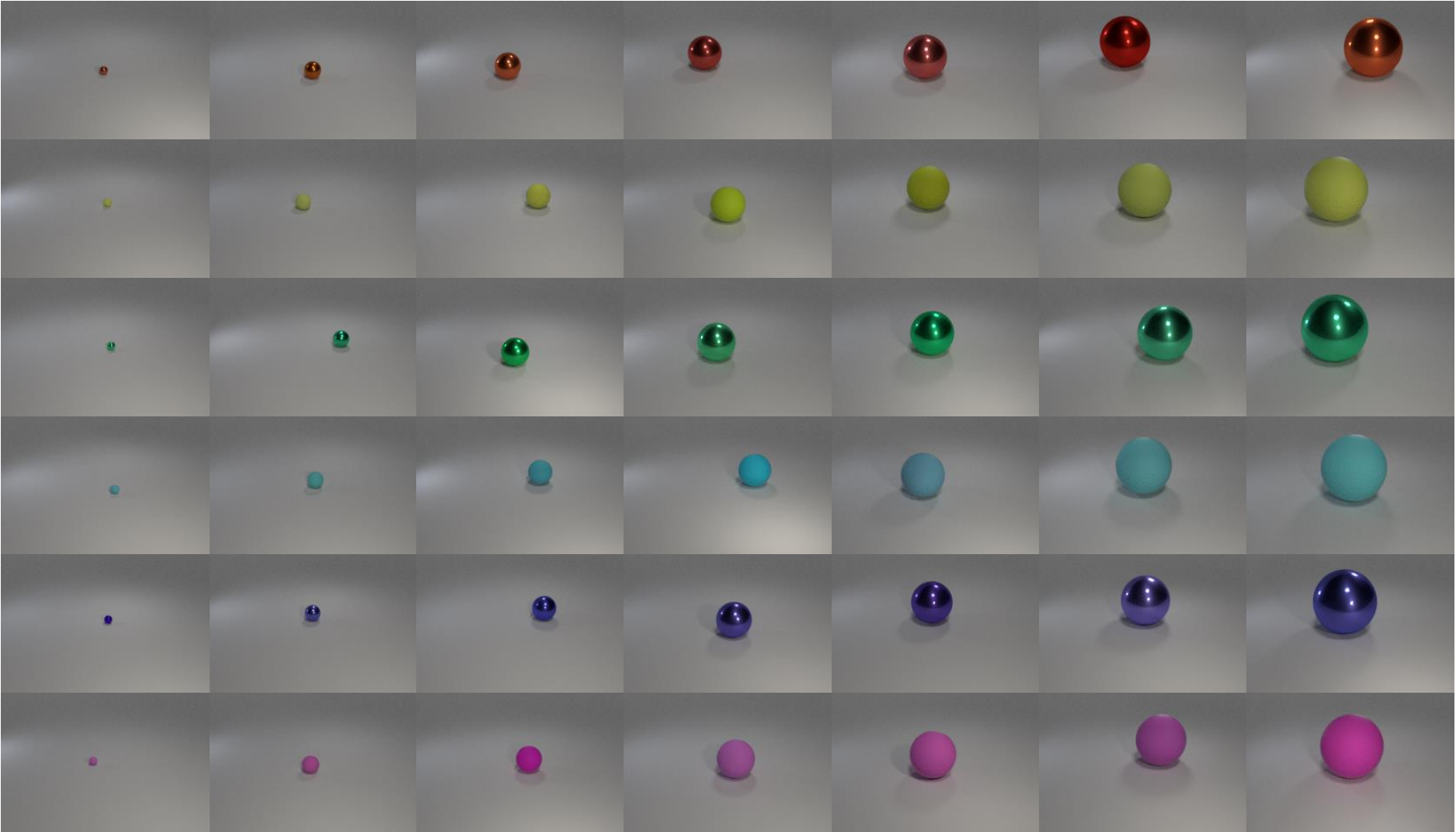}\\
(a) Target domain: sphere images \\\\
\includegraphics[width=0.9\linewidth]{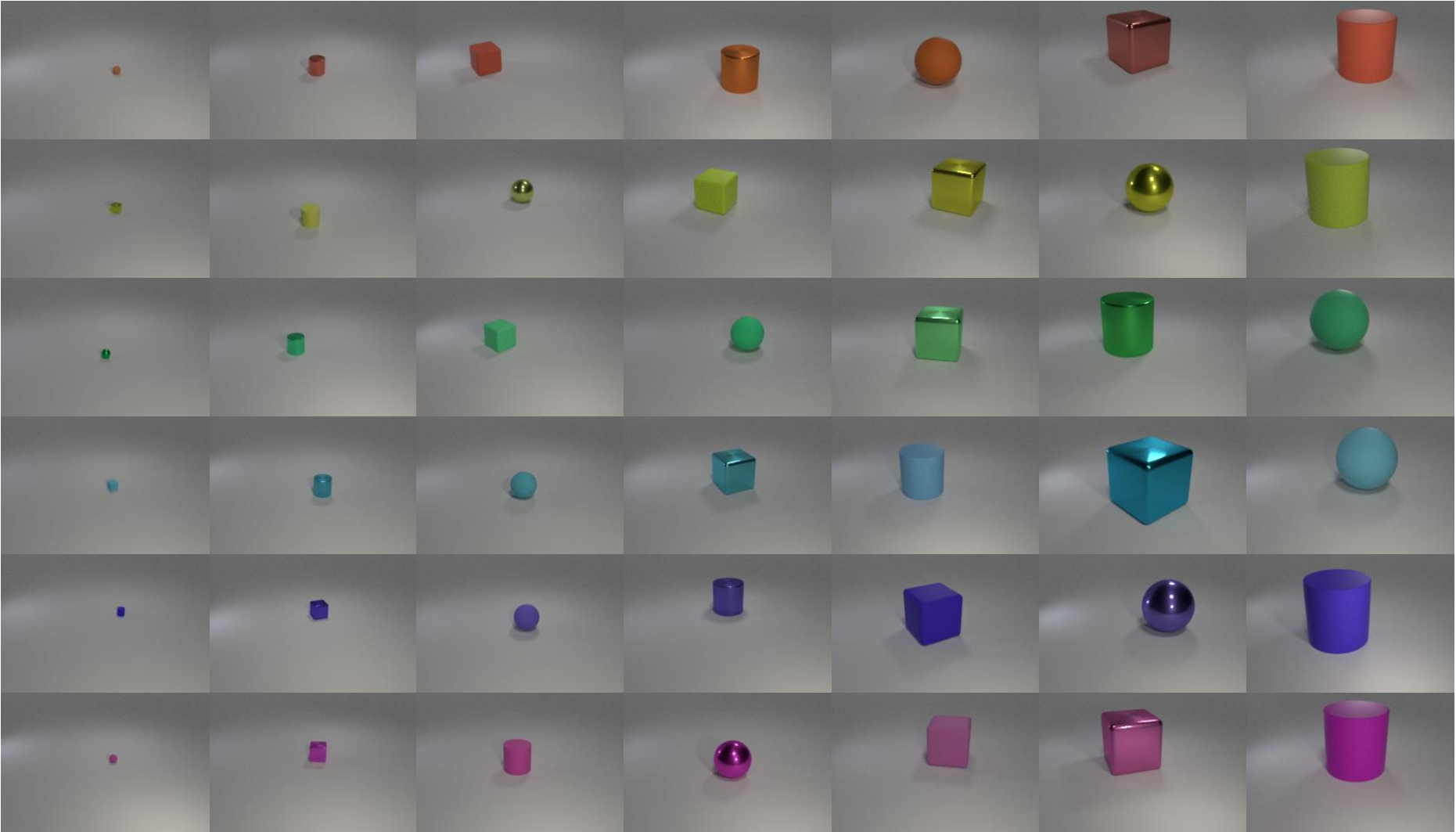}\\
(b) Source domain: sphere, cube, cylinder images \\
 \end{tabular}
\end{center}
 \caption{Image classification (6 classes) for CLEVR.} 
\label{fig:clevr}
\end{figure}



We evaluate ADR in three settings, (1) \textit{perfect simulation} for CLEVR \cite{johnson2017clevr}  ($T$ inside $\Delta$), (2) \textit{imperfect simulation} for Syn2Real \cite{peng2018syn2real}($T$ outside and far from $\Delta$), (3) \textit{average simulation} for VIRAT \cite{sangmin2011virat} ($T$ outside but close to $\Delta$). We perform image classification for the first two settings and object detection for the third setting.

\subsection{Image Classification}
\subsubsection{CLEVR}
We use Blender3D to build a simulator using assets provided by \cite{johnson2017clevr}. The simulator generates images containing exactly one object and labels them into six categories according to the color of the object. The input space $\mathcal{X}$ consists of images with resolution of $480 \times 320$ and the output space $\mathcal{Y}$ is \{red, yellow, green, cyan, purple, magenta\}. Here, $\theta \in \Theta$ corresponds to [color, shape, material, size]. Specifically, 6 colors, 3 shapes (sphere, cube, cylinder), 2 materials (rubber, metal) and 6 sizes. Other parameters like lighting, camera pose are randomly sampled.

As a toy target domain, we generate 5000 images consisting only of spheres. Refer Fig. \ref{fig:clevr} for visualizations of target and source domain.

\textbf{ADR Setup}: The policy $\pi_\omega$ consists of $|\text{color}| \times |\text{shape}| \times |\text{material}| \times |\text{size}|=252$ parameters representing a multinomial distribution over $\Theta$, initialized as a uniform distribution. The learner $h$ is implemented as ResNet18 followed by a fully connected layer for classification which is trained end-to-end. The domain classifier $D$ is a small fully connected network accepting 512 dimensional feature vector extracted from conv5 layer of ResNet18 as input.

\begin{figure}[H]
\begin{center}
\includegraphics[width=1\linewidth]{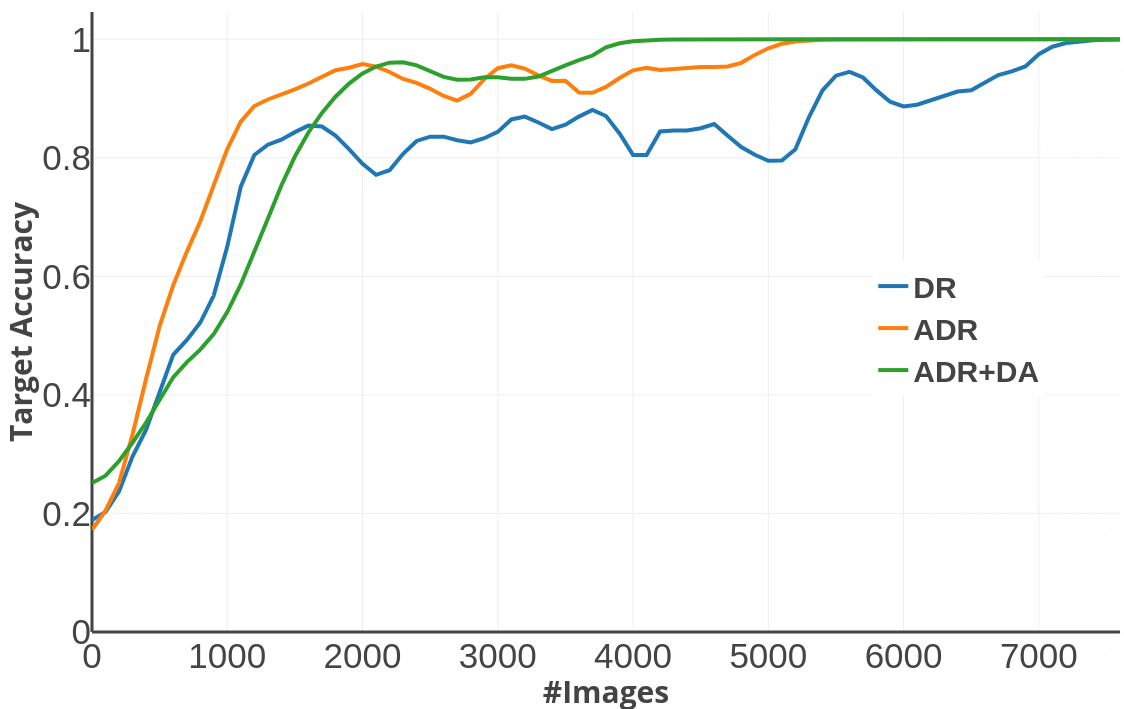}
\end{center}
   \caption{Effect of data size on DR, ADR and ADR+DA, target classification accuracy on CLEVR averaged over 10 independent runs.} 
\label{fig:clevr_target_acc}
\end{figure}

\textbf{Results}: We compare the target classification accuracy of DR, ADR and ADR+DA with the number of training images in Fig. \ref{fig:clevr_target_acc}. Please note that for ADR+DA, we independently generated 1000 unlabeled images from the target domain. ADR eventually learns to generate images containing an object of small size (first and second column in Fig \ref{fig:clevr}). On the other hand, DR keeps on occasionally generating images with large objects, such images are easier for the learner to classify due to a large number of pixels corresponding to the color of object. Interestingly ADR + DA performs the best, the domain classifier learns to discriminate generated images on the basis of the shape being sphere or not. This encourages the policy $\pi_\omega$ to generate images with spheres (images similar to target domain).

\begin{figure}
\begin{center}
\begin{tabular}{c}
\includegraphics[width=\linewidth]{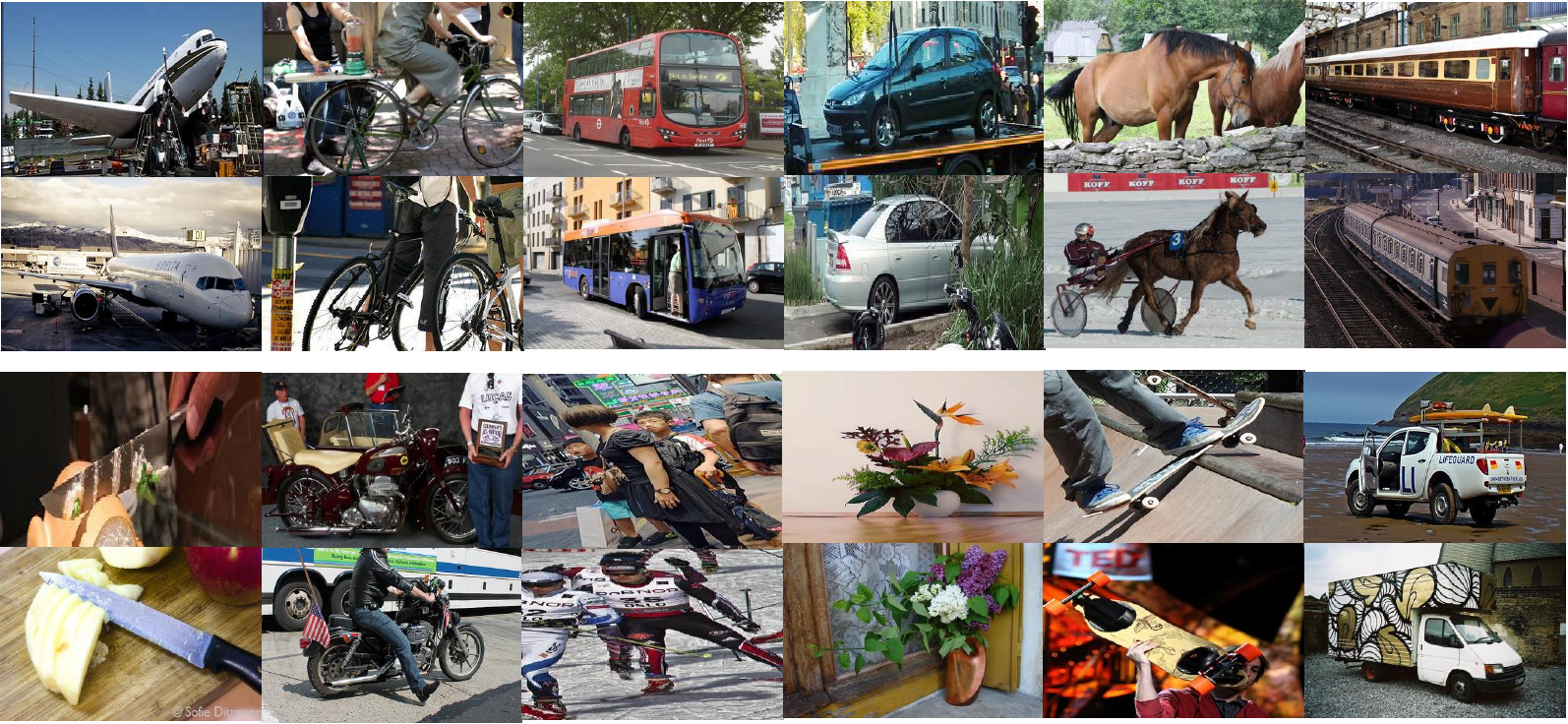}\\
(a) Target domain \\\\
\includegraphics[width=\linewidth]{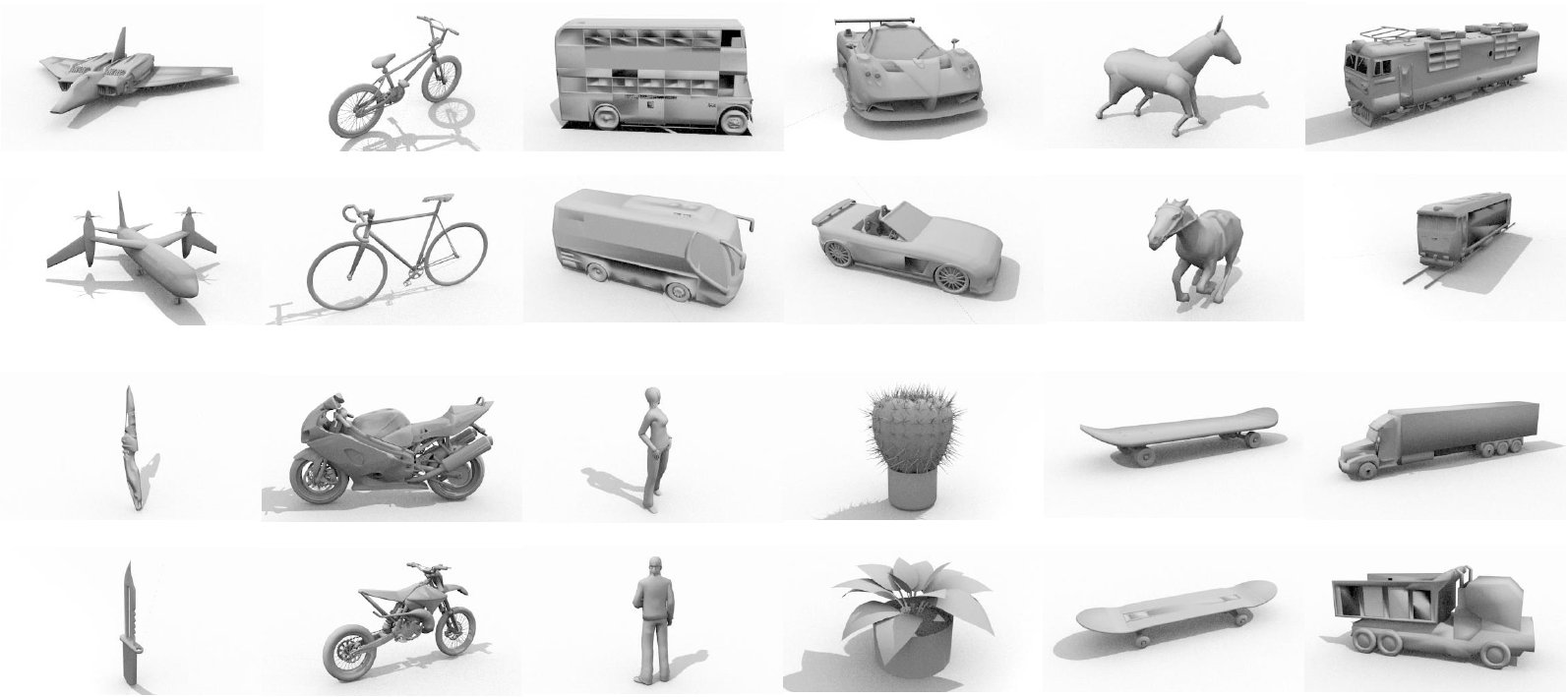}\\
(b) Source domain \\\\
 \end{tabular}
\end{center}
 \caption{Image classification (12 classes) for Syn2Real.} 
\label{fig:syn2real}
\end{figure}


\subsubsection{Syn2Real}
We use 1,907 3D models by \cite{peng2018syn2real} in Blender3D to build a simulator for Syn2Real image classification. In our experiment, we use the validation split of Syn2Real as the target domain. The split contains 55,388 real images from the Microsoft COCO dataset \cite{lin2014coco} for 12 classes. The input space $\mathcal{X}$ consists of images with resolutions $384 \times 216$ and the output space $\mathcal{Y}$ are the 12 categories. Here $\theta \in \Theta$ corresponds to [image class]. Other parameters like camera elevation, lighting and object pose are randomly sampled after quantization. Refer Fig. \ref{fig:syn2real} for visualizations of target and source domain. There are a total of 152,397 possible values of simulation parameters, each corresponds to an unique image in the source domain.



\textbf{ADR Setup}: $\pi_\omega$ consists of $|\text{category of image}|=12$ parameters representing a multinomial distribution over $\Theta$, initialized as a uniform distribution. The learner $h$ is AlexNet \cite{krizhevsky2012imagenet} initialized with weights learned on ImageNet \cite{imagenet_cvpr09}. The hyperparameters are similar to \cite{peng2018syn2real}.

\textbf{Results}: Table \textcolor{red}{1} compares target classification of DR and ADR with varying data size. We observe that ADR focuses on confusing classes like (1) car and bus, (2) bike and motorbike by trading off samples for easier classes like plant and train. Note, we also include the case when every possible image variation (152,397) is used to train the learner $h$ (All-150k). In this case, ADR reduces to hard negative mining over image classes which performs better than the baseline (Source) \cite{peng2018syn2real}. 

We also study the effect of ADR with unlabeled target data and various DA methods like ADA \cite{tzeng2017adversarial}, Deep-CORAL \cite{sun2016return}, DAN \cite{long2015learning}, SE \cite{french2017self} with same setup as \cite{peng2018syn2real} except for ADA, we use a domain classifier with ResNet18 architecture. Refer Table \ref{table:syn2real_da_acc} for per class performance of ADR-150k with various DA methods. Note, we use all the target images without labels for adaptation.

\renewcommand{\arraystretch}{1.5}
\begin{table}
    \centering
    \resizebox{2.8in}{!}{
    \begin{tabular}{cccccc}
    \hline
    \#Images & 10k & 25k & 50k & 100k & All(150k) \\ 
    \hline 
    DR & 8.1 & 10.0 & 13.8 & 23.5 & 28.1 \cite{peng2018syn2real} \\
    \hline 
    ADR & \textbf{15.3} & \textbf{18.6} & \textbf{24.9} & \textbf{31.1} & \textbf{35.9} \\  
    \hline
    \end{tabular}
    }
    \label{tab:syn2real_class_size}
     \vspace{2mm}
\caption{Effect of data size on DR and ADR, target classification accuracy for Syn2Real}
\end{table}
\renewcommand{\arraystretch}{1}

\begin{table*}
  \centering
  \resizebox{5.5in}{!}{
  \begin{tabular}{|l|c|c|c|c|c|c|c|c|c|c|c|c||c||}
  \hline
  DA Method & aero & bike & bus & car & horse & knife & mbke & prsn & plant & skbrd & train & truck & mean \\
  \hline\hline
    All Source \cite{peng2018syn2real}  & 53 & 3 & 50 & 52 & 27 & 14 & 27 & 3 & 26 & 10 & 64 & 4    &  28.1 \\
    ADA  & 68 & 41 & 63 & 34 & 57 & 45 & 74 & 30 & 57 & 24 & 63 & 15 & 47.6  \\
    D-CORAL \cite{peng2018syn2real} & 76 & 31 & 60 & 35 & 45 & 48 & 55 & 28 & 56 & 28& 60 & 19 & 45.5\\
    DAN \cite{peng2018syn2real} & 71 & 47 & 67 & 31 & 61 & 49 & 72 & 36 & 64 & 28 & 70 & 19 & 51.6\\
    SE \cite{peng2018syn2real}  & 97 & 87 & 84 & 64 & 95 & 96 & 92 & 82 & 96 & 92 & 87 & 54 & 85.5 \\
    
    \hline
    ADR     & 49 & 12 & 52 & 56 & 38 & 25 & 31 & 14 & 34 & 32 & 59 & 29  & 35.9  \\
    ADA + ADR  & 73 & 50 & 60 & 38 & 59 & 51 & 79 & 37 & 60 & 41 & 69 & 35 & 54.3  \\
    D-CORAL + ADR  & 78 & 56 & 71 & 48 & 64 & 59 & 77 & 45 & 68 & 49 & 70 & 55 & 61.6\\
    DAN + ADR & 87 & 60 & 73 & 40 & 59 & 56 & 68 & 43 & 72 & 39 & 68 & 51 & 59.7\\
    SE + ADR  & 94 & 85 & 88 & 72 & 89 & 93 & 91 & 88 & 93 & 86 & 84 & 75 & 86.4 \\
    \hline
    Real \cite{peng2018syn2real} & 94 & 83 & 83 & 86 & 93 & 91 & 90 & 86 & 94 & 88 & 87 & 65 & 87.2\\
    \hline
  \end{tabular}
  }
  \vspace{2mm}
  \caption{Effect of unlabeled target data with ADR on Syn2Real}
  \label{table:syn2real_da_acc}
\end{table*}

\subsection{Object Detection}


\begin{figure}
\begin{center}
\begin{tabular}{c}
 \includegraphics[width=0.9\linewidth]{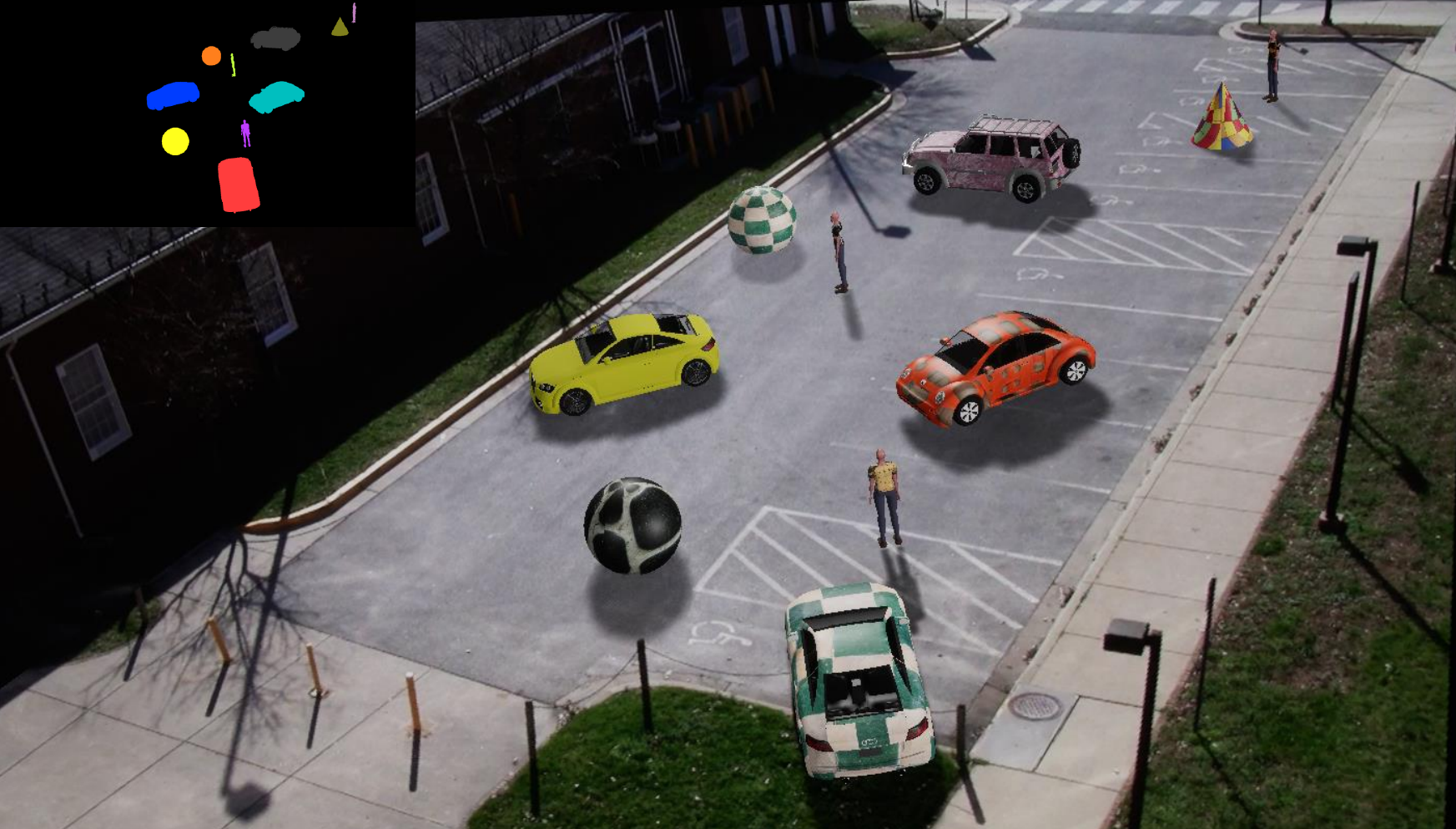}\\
 (a) Domain Randomization \\
 \includegraphics[width=0.9\linewidth]{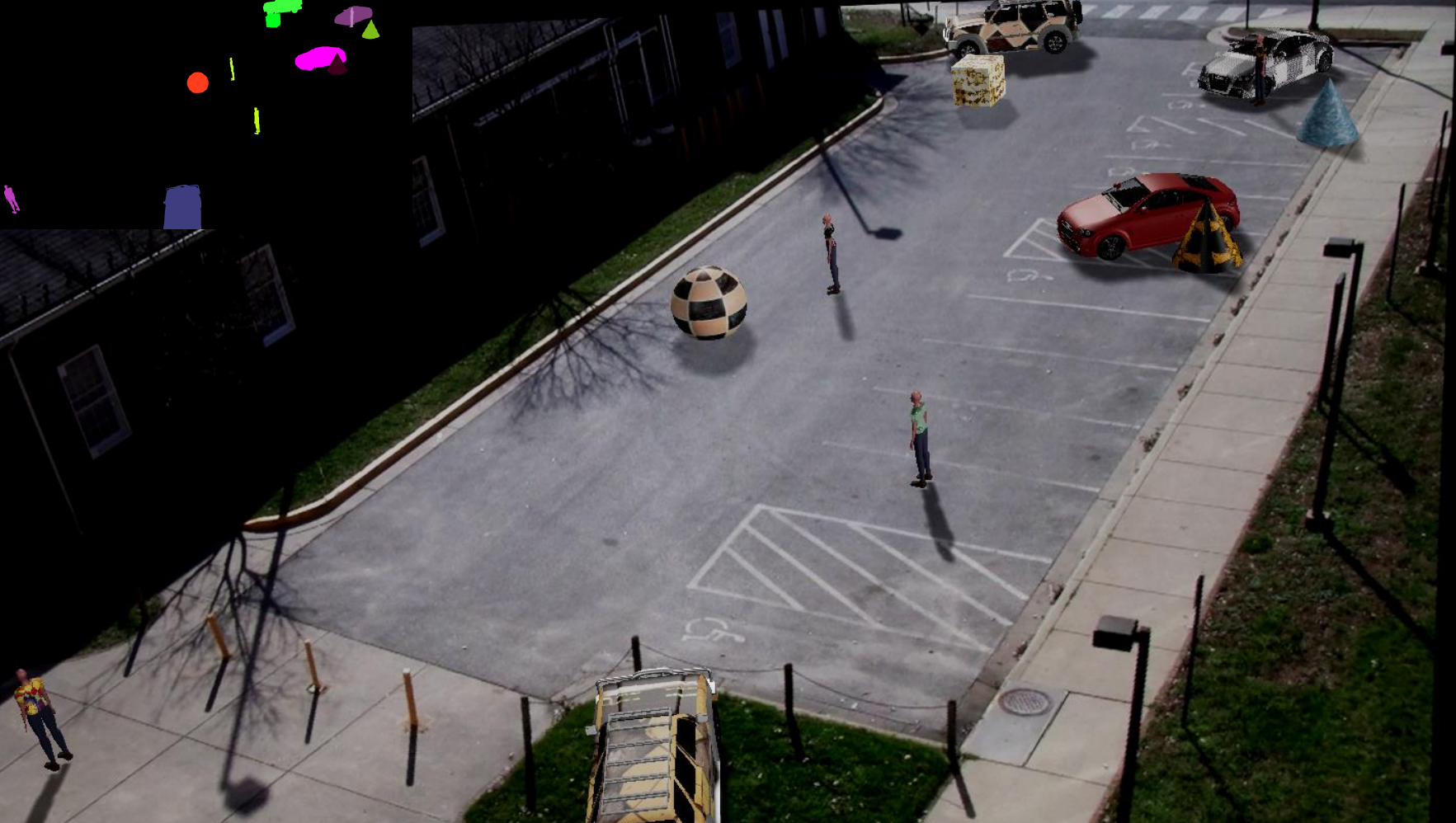} \\
(b) Adversarial Domain Randomization
 \end{tabular}
\end{center}
  \caption{A comparison of source data for VIRAT from DR and ADR along with ground truth instance segmentation map.}
\label{fig:virat_scene1}
\end{figure}

We use the Unreal Engine 4 based simulator by \cite{khirodkar2018domain} for the VIRAT dataset \cite{sangmin2011virat}. VIRAT contains surveillance videos of parking lots. The simulator models the parking lot and the surveillance camera in a 3D world. The randomization process uses a texture bank of 100 textures with 10 cars, 5 person models and geometric distractors (cubes, cones, spheres) along with varying lighting conditions, contrast and brightness. In our experiment, we use 50,000 images from two scenes of the dataset as the target domain. The input space $\mathcal{X}$ consists of images with resolution $1920 \times 1080$ and the output space $\mathcal{Y}$ is the space of car bounding boxes (atmost 20) present in the image. Here $\theta \in \Theta$ is a list of object attributes in the image. These attributes specify the location and type of the object in the image. Fig. \ref{fig:ADR_intro}, \ref{fig:virat_scene1} shows labeled samples from source domain.

\begin{figure}
\begin{center}
\begin{tabular}{c}
\includegraphics[width=0.8\linewidth]{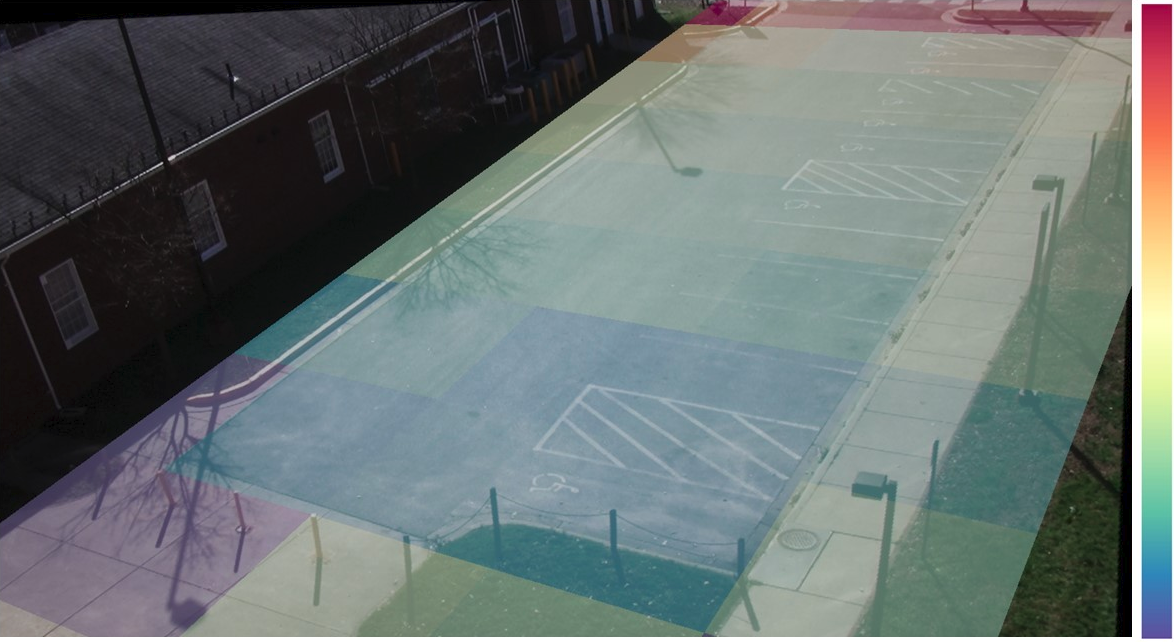}\\
\includegraphics[width=0.8\linewidth]{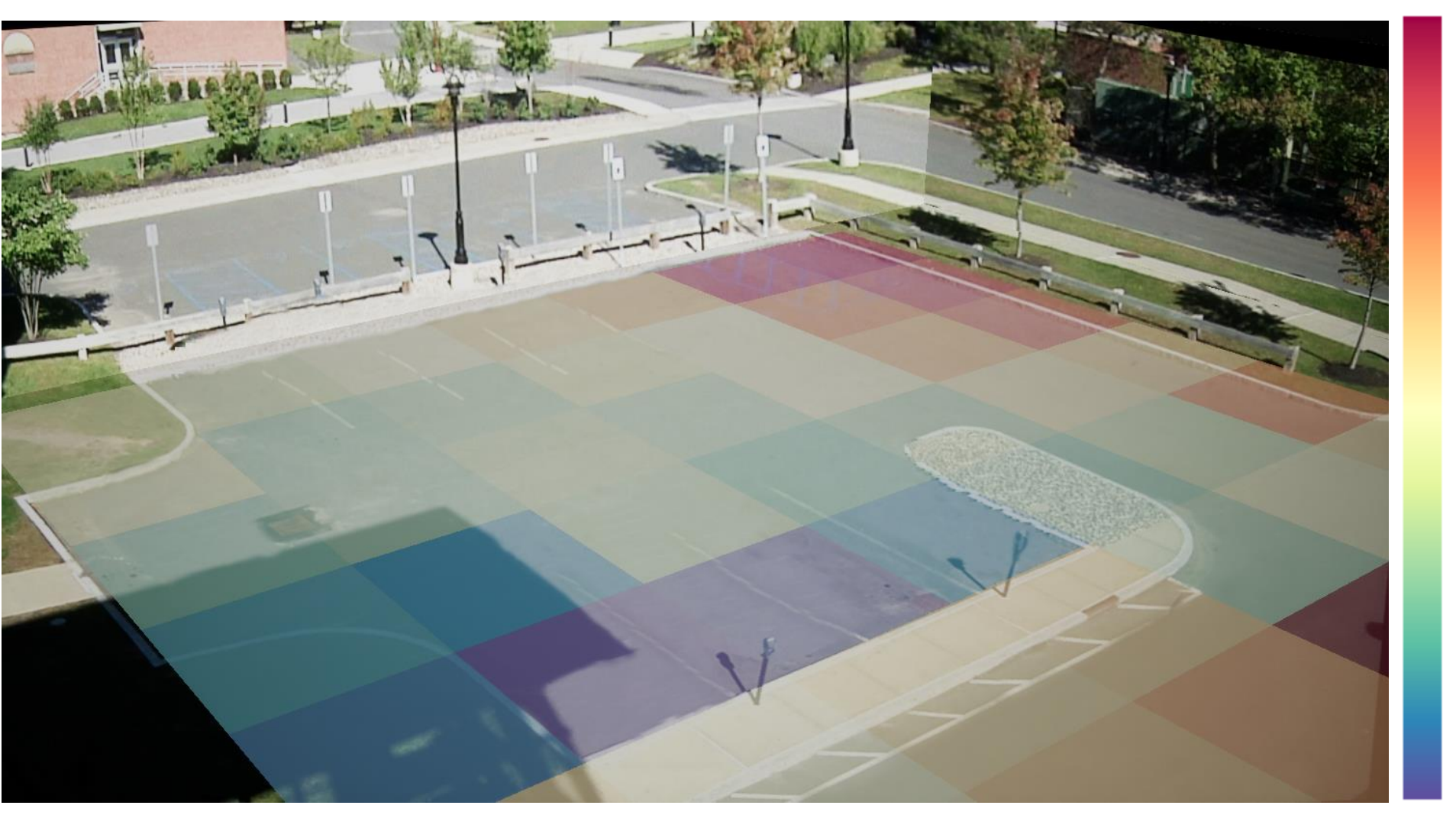}
 \end{tabular}
\end{center}
  \caption{Object spawn probability ($\pi_\omega$) visualized as a heat-map. The warmer colors indicate higher probability which correspond to small/truncated/occluded objects in the image.} 
\label{fig:virat_spawn}
\end{figure}

\textbf{ADR Setup:}
We divide the scene ground plane into 45 rectangular cells (refer Fig. \ref{fig:virat_spawn}). In each cell, we place three kinds of objects (car, person or distractor). The policy $\pi_\omega$ consists of $|\text{cells}| \times |\text{types of objects}| = 135$ parameters representing the object spawn probability \ie the policy $\pi_\omega$ decides which object is placed where in the scene. To include variable number of objects in the image, we randomly sample $n$ = number of objects ($2 \leq n \leq 12)$ and invoke $\pi_\omega$ $n$ times. Other parameters like lighting, texture, car model, pose, size of the object are randomly sampled. The reward for policy $\pi_\omega$ is computed per cell and is negative of the IoU of the bounding box predicted by the learner $h$. 

The learner $h$ is implemented as Faster-RCNN \cite{ren2015faster} with RoI-Align and ResNet101 with feature pyramid network \cite{lin2017feature} architecture as the backbone for all our evaluations.

\renewcommand{\arraystretch}{1.5}
\begin{table}[H]
    \centering
    \resizebox{2.8in}{!}{
    \begin{tabular}{cccccc}
    \hline
    \#Images & 1k & 10k & 25k & 50k & 100k \\ 
    \hline 
    DR & 20.6 & 32.1 & 43.7 & 54.9 & 75.8 \\ 
    \hline 
    ADR & \textbf{31.4} & \textbf{43.8} & \textbf{56.0} & \textbf{78.2} & \textbf{88.6} \\ 
    \hline
    \end{tabular}
    }
    \vspace{2mm}
    \caption{Effect of data size on DR and ADR, Faster-RCNN's AP at 0.7 IoU on VIRAT.}
    \label{tab:obj_det_size}
\end{table}
\renewcommand{\arraystretch}{1}


\begin{table}
    \centering
    \resizebox{2in}{!}{
    \begin{tabular}{|l|c|c|}
    \hline
    Method  & AP @ 0.7 \\
    \hline
    COCO & 80.2  \\
    ADA & 84.7  \\
    ADR-100k + ADA & \textbf{93.6} \\
    \hline
    Real & 98.1\\
    \hline
    \end{tabular}
    }
    \vspace{2mm}
    \caption{Effect of unlabeled target data with ADR on VIRAT. }
    \label{table:virat_da}
\end{table}

\begin{figure}
\begin{center}
\begin{tabular}{c}
 \includegraphics[width=0.8\linewidth]{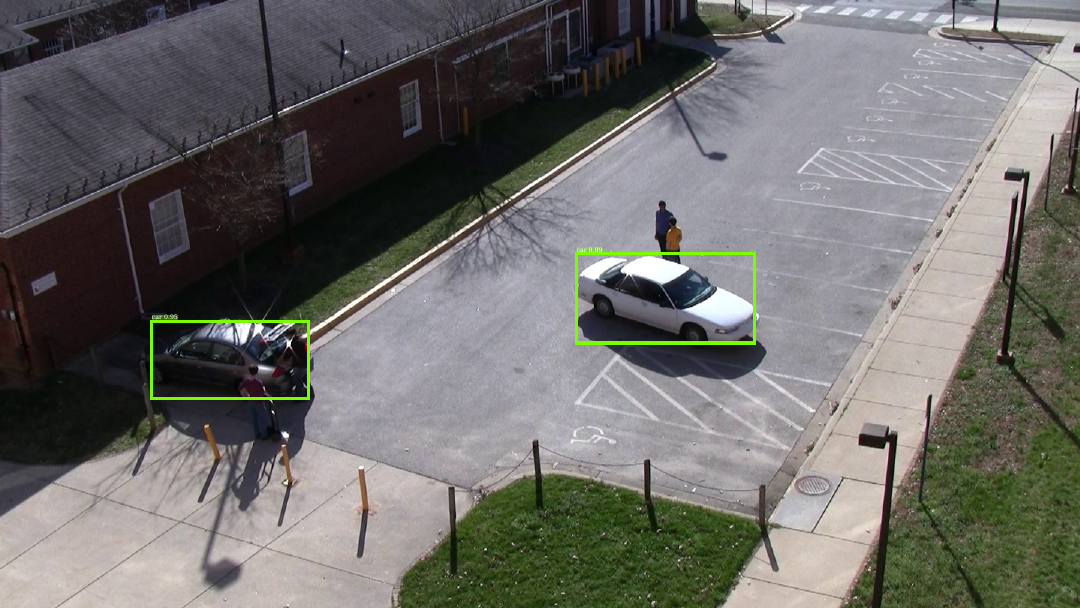}\\
  \includegraphics[width=0.8\linewidth]{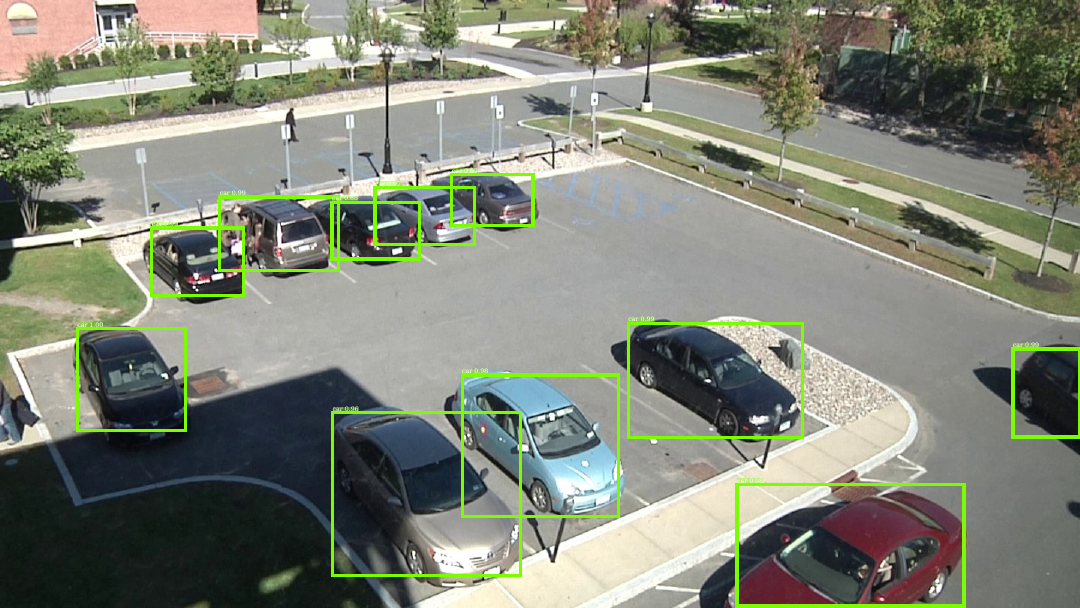}
 \end{tabular}
\end{center}
   \caption{Faster-RCNN trained on 100k images generated by ADR}
\label{fig:ADR_det}
\end{figure}

\textbf{Results}:
Table \ref{tab:obj_det_size} compares performance of DR and ADR on target data along with the size of synthetic data. ADR outperforms DR consistently by generating informative samples for object detection containing object occlusions and truncations. We compare data samples generated by DR and ADR in Fig. \ref{fig:ADR_intro}, \ref{fig:virat_scene1} along with a visualization of $\pi_\omega$ learned by ADR in Fig. \ref{fig:virat_spawn}, $\pi_\omega$ is shown as a heat-map (warmer colors indicate higher object spawn probability). ADR learns to place objects far from the camera, thus making them difficult for the learner to detect. Fig. \ref{fig:ADR_det} shows examples of car detection from Faster-RCNN trained on data generated by ADR, affirming that our method performs well under severe truncations and occlusions.





We also evaluate ADR with 5,000 unlabeled target images (not in the test set). Refer Table \ref{table:virat_da} for comparison of (1) Faster-RCNN trained on Microsoft-COCO dataset (COCO), (2) ADA \cite{tzeng2017adversarial} with ResNet18 as domain classifier, (3) ADR + ADA with 100k source images and (4) Faster-RCNN trained on target images from VIRAT (Real). Using unlabeled data with synthetic data boosts performance from (for DR 75.8 to 84.7, for ADR 88.6 to 93.6). However, the combination of labeled synthetic data and unlabeled real data still performs worse than labeled real data. We provide more analysis in supplementary material.